%% file: main.tex
\documentclass[10pt,twocolumn,letterpaper]{article}

\usepackage[pagenumbers]{iccv} % To force page numbers, e.g. for an arXiv version
\usepackage{graphicx}
\usepackage{amsmath}
\usepackage{amssymb}
\usepackage{booktabs}
\usepackage{animate}
\usepackage{enumitem}
\usepackage{multirow}
\usepackage{diagbox}

\usepackage{graphicx}
\usepackage{amsmath}
\usepackage{amssymb}
\usepackage{booktabs}
\usepackage{float}
\usepackage{diagbox}
\usepackage{multirow}
\usepackage{enumitem}
\usepackage{tabularx}
\usepackage{microtype}
\usepackage{soul}

\usepackage{times}
\usepackage[T1]{fontenc}

\pdfmapfile{+times.map}

\usepackage[numbers,sort&compress]{natbib}

\makeatletter
\@namedef{ver@everyshi.sty}{}
\makeatother
\usepackage{tikz}
\usepackage{makecell}
\usepackage{pgfplots}
\usetikzlibrary{spy,calc}

\definecolor{iccvblue}{rgb}{0.21,0.49,0.74}
\usepackage[pagebackref,breaklinks,colorlinks,allcolors=iccvblue]{hyperref}

\usepackage[capitalize]{cleveref}
\crefname{section}{Sec.}{Secs.}
\Crefname{section}{Section}{Sections}
\Crefname{table}{Table}{Tables}
\crefname{table}{Tab.}{Tabs.}

\def\paperID{1551} % *** Enter the Paper ID here
\def\confName{ICCV}
\def\confYear{2025}

\definecolor{Gray}{gray}{0.5}
\definecolor{LightCyan}{rgb}{0.88,1,1}

\newcolumntype{a}{>{\columncolor{Gray}}c}
\newcolumntype{b}{>{\columncolor{white}}c}

\begin{document}


\title{Stroke2Sketch: Harnessing Stroke Attributes for Training-Free Sketch Generation} % *** Enter the paper title here

\author{Rui Yang$^{1,3}$, Huining Li$^{4, 5}$, Yiyi Long$^{2, 5}$, Xiaojun Wu$^{3, }$\thanks{Corresponding authors: Xiaojun Wu (xjwu@snnu.edu.cn) and Shengfeng He (shengfenghe@smu.edu.sg)}~, Shengfeng He$^{5, *}$\\
$^{1}$Huaqiao University\quad$^{2}$South China University of Technology\quad$^{3}$Shaanxi Normal University\\
$^{4}$Beijing University of Aeronautics and Astronautics\quad$^{5}$Singapore Management University\\
% {\tt\small }
}

\teaser{
\centering
\includegraphics[width=0.95\textwidth]{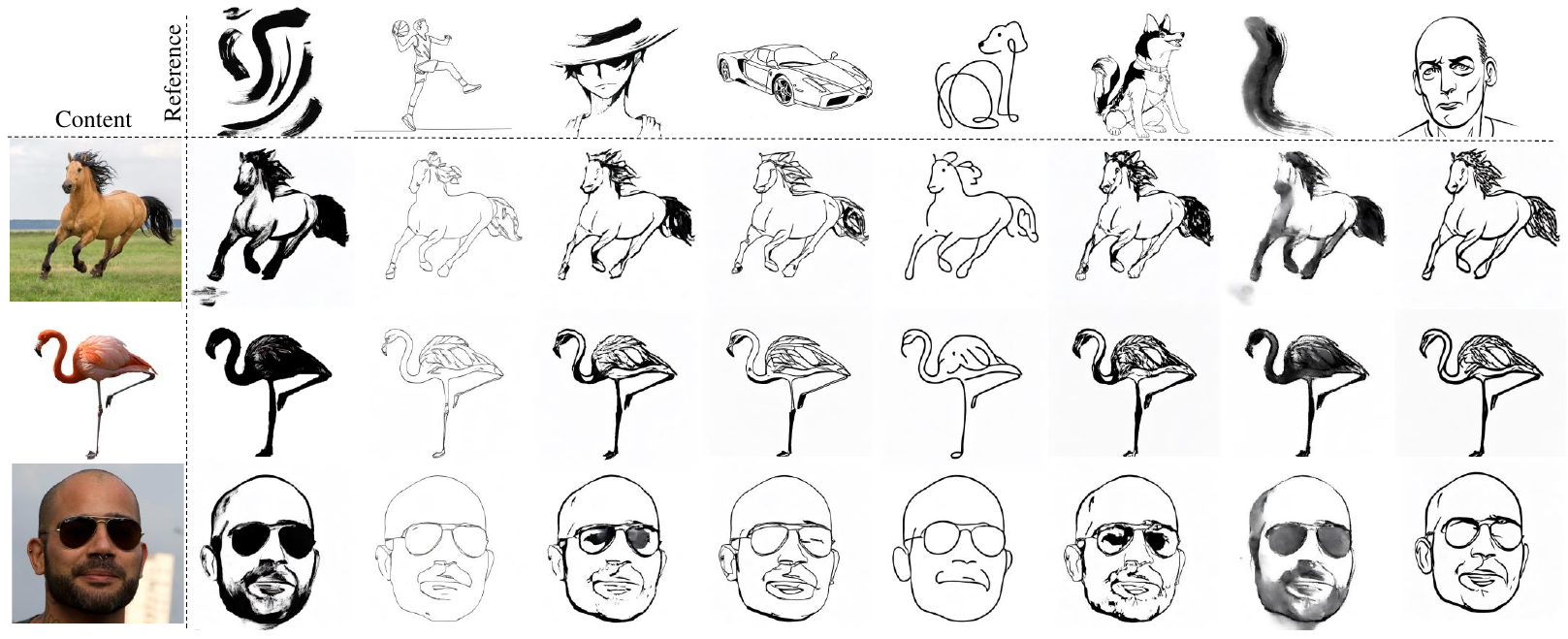}
\vspace{-3mm}\caption{We propose Stroke2Sketch, a framework that accurately transfers stroke attributes from a reference sketch to a content image while preserving structure and style fidelity. The top row shows reference sketches, the leftmost column displays content images, and the central and right columns illustrate our method’s precise content preservation and expressive stroke transfer.}\vspace{-5mm}
\label{fig1}
}
\maketitle

\input{./sec/0_abstract.tex}

\input{./sec/1_intro.tex}

\input{./sec/2_related}
\input{./sec/3_method}

\input{./sec/4_experiment}
\input{./sec/5_discussion}
\input{./sec/suppl}

%%%%%%%%% REFERENCES
{\small
\bibliographystyle{ieeenat_fullname}
\bibliography{./main}
}

\end{document}

%% file: sec/0_abstract.tex
\begin{abstract}%\vspace{-0.5mm}

Generating sketches guided by reference styles requires precise transfer of stroke attributes, such as line thickness, deformation, and texture sparsity, while preserving semantic structure and content fidelity. To this end, we propose Stroke2Sketch, a novel training-free framework that introduces cross-image stroke attention, a mechanism embedded within self-attention layers to establish fine-grained semantic correspondences and enable accurate stroke attribute transfer. This allows our method to adaptively integrate reference stroke characteristics into content images while maintaining structural integrity. Additionally, we develop adaptive contrast enhancement and semantic-focused attention to reinforce content preservation and foreground emphasis. Stroke2Sketch effectively synthesizes stylistically faithful sketches that closely resemble handcrafted results, outperforming existing methods in expressive stroke control and semantic coherence.  Codes are available at \href{https://github.com/rane7/Stroke2Sketch}{https://github.com/rane7/Stroke2Sketch}.
\end{abstract}\vspace{-3mm}

%% file: sec/1_intro.tex
\begin{figure*}
    \centering
    \includegraphics[width=\linewidth]{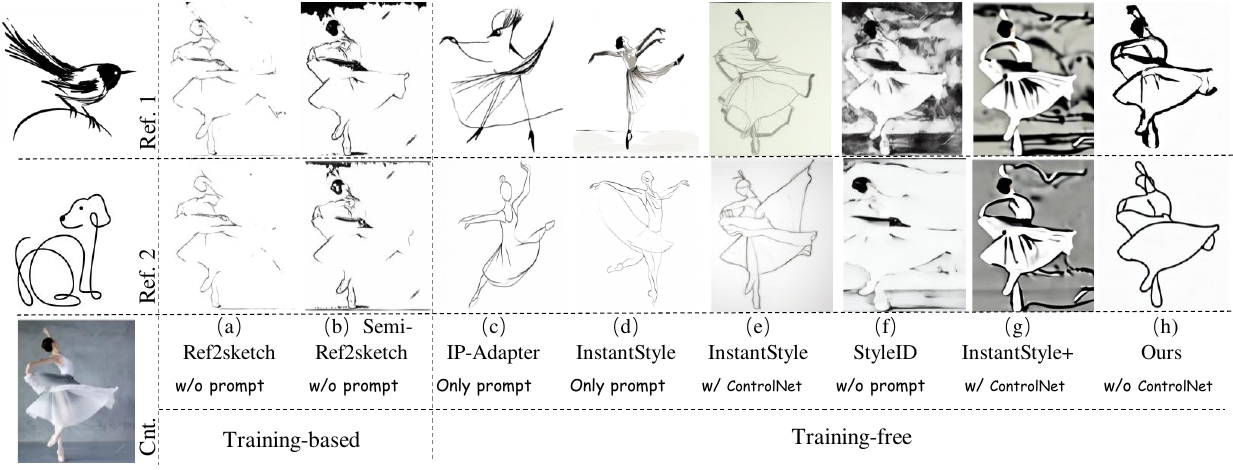}
    \vspace{-5mm}\caption{(a) and (b) show results from training-based methods (Ref2Sketch and Semi-Ref2Sketch), which struggle with unseen styles, leading to poor content alignment. (c) and (d) illustrate IP-Adapter and InstantStyle results, retaining style but lacking content alignment. (e) to (g) show ControlNet-based methods, preserving content but failing to match reference styles. (h) shows our method’s output, achieving superior content preservation and style alignment without using ControlNet. Detailed prompts are shown in the supplementary materials.}
    \label{fig2}
    \vspace{-4mm}
\end{figure*}
\label{sec1}

\section{Introduction}
\label{sec:1}

Generating stylized sketches from content images using reference stroke patterns presents a key challenge at the intersection of artistic rendering and semantic-aware synthesis. Traditional sketch algorithms~\cite{nakano2019neural,10.1145/2897824.2925972} generate procedural line drawings, while vector-based methods~\cite{DBLP:conf/iccv/LiuFHK21,10319299,vinker2022clipasso} produce clean parametric strokes. However, these methods lack the adaptability to transfer diverse artistic styles from exemplar sketches due to their rigid, data-agnostic architectures. Unlike human artists, who strategically vary stroke attributes such as thickness, curvature, and density to emphasize key semantic features while maintaining content fidelity, existing approaches struggle to capture this nuanced interplay between stroke semantics and structure.

Recent learning-based methods~\cite{ref2sketch,semi2sketch,chan2022learning} attempt to address this limitation by training on clustered sketch styles, yet as shown in Fig.~\ref{fig2}(a-b), they fail to generalize to unseen stroke patterns due to catastrophic forgetting. Meanwhile, diffusion-based stylization techniques~\cite{wang2024instantstyle,ye2023ip,yu2024beyond,li2025personamagic,xu2024dreamanime} excel in texture transfer but struggle with structural integrity due to content leakage in cross-attention layers (Fig.~\ref{fig2}(c-d)). While new conditioning mechanisms~\cite{liu2024drag,jiang2023diffuse3d,wu2022make} attempt to enhance structural control, they often introduce style distortion or require dense user inputs. Hybrid approaches like ControlNet~\cite{zhang2023adding} enforce structural priors but sacrifice stylistic flexibility, leading to overly rigid outputs (Fig.~\ref{fig2}(e)). Progressive stroke-based methods~\cite{rout2024rbmodulation} introduce semantic incoherence by applying uniform strokes across the image (Fig.~\ref{fig2}(g)).

We identify three fundamental challenges in reference-based sketch synthesis:\textit{(i) Semantic-aware stroke transfer.} Effective style adaptation requires precise mapping of reference stroke attributes (e.g., tapered lines, cross-hatching) to semantically relevant content regions. \textit{(ii) Foreground prioritization.} Artists naturally emphasize foreground elements using varied stroke density and complexity while simplifying backgrounds, yet existing methods apply uniform stylization, disrupting this compositional balance\cite{rout2024rbmodulation}. \textit{(iii) Content-style equilibrium.} Since sketches encode content through linework, balancing structural preservation with style transfer is critical. Techniques such as CLIP-space style subtraction~\cite{wang2024instantstyle} fail to maintain this balance, as even minor content leakage distorts key edges (Fig.~\ref{fig2}(c-d)).

To address these challenges, we propose Stroke2Sketch, a framework that enables precise stroke attribute transfer while maintaining content fidelity. Our key insight is that stroke properties, like line thickness, curvature, and texture sparsity, are inherently encoded within the self-attention and cross-attention relationships of pretrained diffusion models. By dynamically aligning these attention patterns between content and reference features, we achieve effective style transfer without structural degradation. Stroke2Sketch integrates three novel components tailored to each identified challenge:

\noindent\textit{(1) Cross-image stroke attention} tackles the challenge of semantic-aware stroke transfer by facilitating stroke attribute exchange through key-value swapping in diffusion layers. Instead of directly blending features~\cite{wang2024instantstyle}, we leverage attention blocks to transplant stroke characteristics onto content structures, preventing the entanglement of style and geometry. This approach ensures accurate stroke mapping without disrupting semantic coherence, as shown in Fig.~\ref{fig2}(h).

\noindent\textit{(2) Directive Attention Module} ensures that stroke transfer remains compositionally balanced. Background textures often introduce conflicting patterns that dilute the intended style. We mitigate this by clustering self-attention maps to isolate foreground objects, then restricting cross-image attention to these regions. This mimics how artists prioritize focal elements while simplifying less critical areas, enhancing both style consistency and perceptual quality.

 \noindent\textit{(3) Semantic Preservation Module} addresses content-style equilibrium by injecting content contours as positional queries during early denoising. This hybridizes the precision of edge detectors with the flexibility of text-driven generation, allowing structural guidance without rigid constraints. Unlike ControlNet, which enforces strict boundaries, our approach treats edges as ``soft constraints'' that evolve into stylized strokes, preserving both structure and artistic abstraction.

As validated in Fig.~\ref{fig2}(h) and Fig.~\ref{fig6}, Stroke2Sketch achieves state-of-the-art performance across diverse sketch styles. Experimental results show that our method outperforms adapter-based and ControlNet methods in both style alignment (87\% user preference) and content preservation (92\% accuracy in line correspondence tests). Importantly, it achieves these results without dataset-specific training or architectural modifications, demonstrating that pretrained diffusion models can master sketch stylization when guided by principled feature interactions.

%% file: sec/2_related.tex
\section{Related Work}
\label{sec:2}

%-------------------------------------------------------------------------
\subsection{Photo-Sketch Synthesis}
Generating a sketch from a content image based on a reference style parallels edge detection, as both tasks emphasize prominent visual features. Edge detection methods~\cite{canny1986computational, xie2015hed, soria2023tiny, zhou2024muge, poma2020dense, su2021pixel}, which detect sharp changes in color or brightness, form the foundation of sketch extraction but are limited to a single style and often produce artifacts, like scattered dots or broken lines. High-quality sketch generation requires more than edge detection; it demands line style, texture sparsity, and semantic abstraction to achieve artist-level results.

Learning-based approaches have improved sketch realism by enhancing boundary detection and rendering distinct line styles. For example, Chan et al.~\cite{chan2022learning} incorporated depth and semantic information for better sketch quality, while Ref2Sketch~\cite{ref2sketch} and Semi-Ref2Sketch~\cite{semi2sketch} leverage paired and semi-supervised training for stylized sketch extraction. However, these methods rely on large sketch datasets, which are challenging to obtain, limiting model robustness.

Another research area, stroke-based rendering, focuses on manipulating strokes and contours for sketching. Methods like CLIPDraw~\cite{frans2022clipdraw} and CLIPasso~\cite{vinker2022clipasso} employ Bezier curves to create abstract sketches with high-level semantic simplification. StrokeAggregator~\cite{liu2018strokeaggregator} and StripMaker~\cite{10.1145/3592130} refine vector sketches, achieving quality comparable to artist drawings. Other methods include semantic concept-to-sketch methods~\cite{cao2019ai, zang2023self,das2020beziersketch,hu2024scaleadaptive}. However, these methods assume uniform style, while real-world sketches vary widely.

Building on these insights, our method combines stroke attributes, semantic abstraction, and expression to better align generated sketches with diverse reference styles.

%-------------------------------------------------------------------------

\subsection{High-Semantic Style Transfer}
Generating sketches that adhere to both content and reference style is a specialized style transfer task. Advances in diffusion models have propelled style transfer through self- and cross-attention mechanisms, which preserve spatial layout and stylized content. Techniques such as Prompt-to-Prompt~\cite{hertz2022prompt} and P+~\cite{voynov2023p+} show how attention mechanisms can maintain structural coherence while enabling flexible style and semantic control.

Cross-Image Attention (CIA)~\cite{cia} and methods like Swapping Self-Attention~\cite{jeong2024visualswapp} and StyleAligned~\cite{hertz2024stylealigned} reveal that style features are best retained during upsampling stages, even though content leakage may occur in bottleneck phases. IP-Adapter achieves style transfer via dual cross-attention with text and image, though it can weaken text control and lead to leakage.

InstantStyle~\cite{wang2024instantstyle} addresses leakage by subtracting features in the same feature space but often requires external constraints like ControlNet~\cite{zhang2023adding} for image-to-image generation, which can dilute style fidelity. Further developments like InstantStyle-plus~\cite{insp} and RB-Modulation~\cite{rout2024rbmodulation} incorporate CSD~\cite{somepalli2024measuring} to better align styles during generation, but limitations remain in sketch generation. Prompt-to-style transfer, however, demonstrates pretrained models’ strong semantic alignment capabilities, providing valuable priors for consistent, high-level sketch synthesis.

%-------------------------------------------------------------------------

%% file: sec/3_method.tex
\begin{figure*}[h]
  \includegraphics[width=\linewidth]{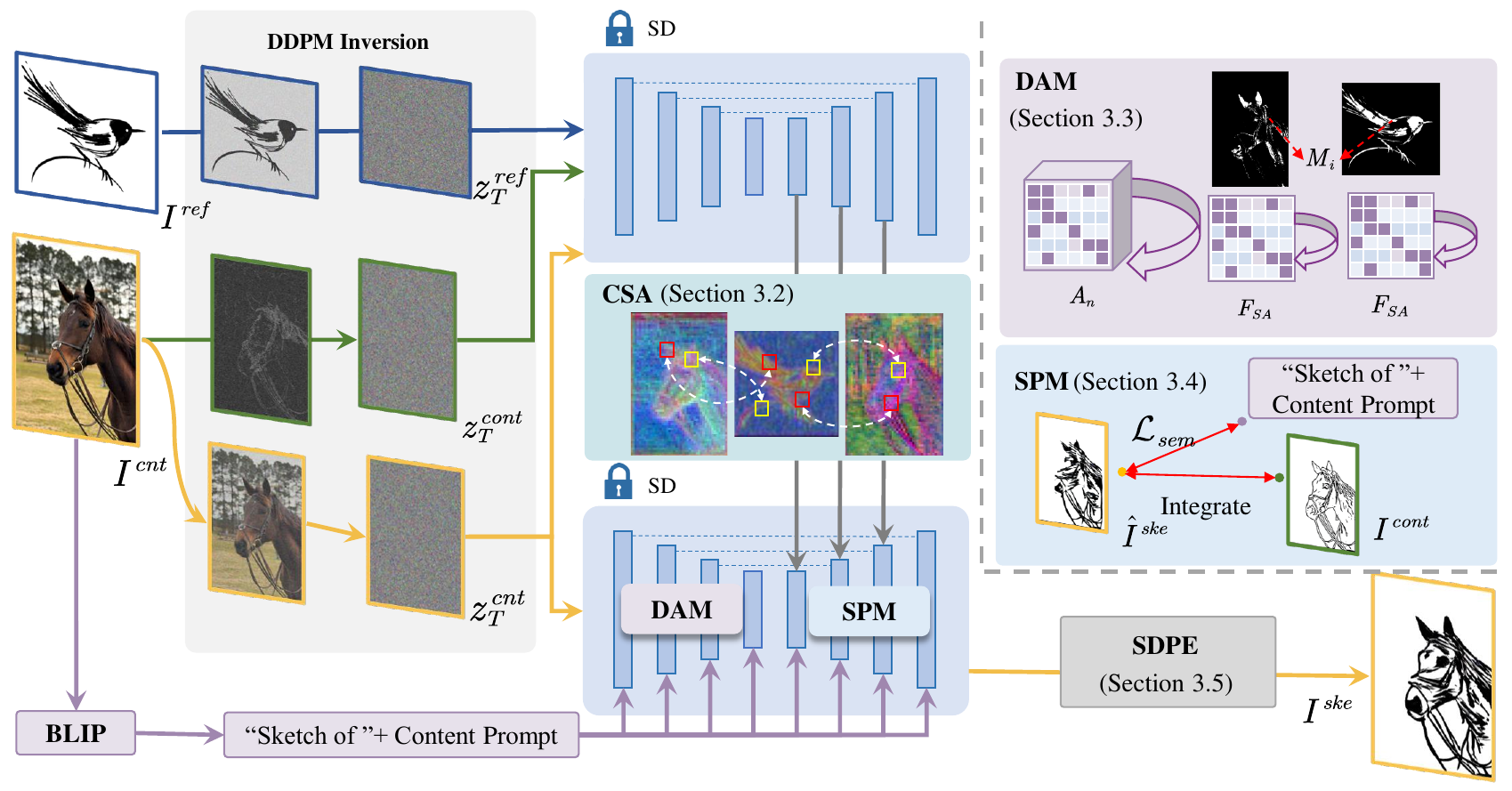}
  \vspace{-3mm}\caption{The Stroke2Sketch architecture. The content image \( I^{\text{cnt}} \) undergoes contour detection, generating feature representations \( z^{\text{cont}} \) and \( z^{\text{cnt}} \), while the reference sketch \( I^{\text{ref}} \) is inverted to produce \( z^{\text{ref}} \). The Directive Attention Module (DAM) aligns high-level semantic features between the content and reference features by emphasizing cross-image semantic correspondences \( A_{n} \). Self-attention maps $F_{SA}$ are aggregated and clustered to produce segmentation masks $M_{i}$, which help distinguish foreground from background regions. Cross-image Stroke Attention (CSA) transfers stroke attributes, and the Semantic Preservation Module (SPM) enforces semantic alignment and stroke fidelity in the generated sketch \( \hat{I}^{\text{ske}} \) via loss \( \mathcal{L}_{\text{sem}} \) and contour-based structural integration. Lastly, Stroke Detail Propagation Enhancement (SDPE) refines details, resulting in the final output sketch \( I^{\text{ske}} \). SD represents the pre-trained diffusion model.}
  \label{fig3}
\end{figure*}

\section{Stroke2Sketch}
\label{sec:3}

Given a content image \( I^{cnt} \in \mathbb{R}^{H^{cnt} \times W^{cnt} \times 3} \) and a reference sketch image \( I^{ref} \in \mathbb{R}^{H^{ref} \times W^{ref} \times 3} \), our task is to generate a sketch \( I^{ske} \) that aligns with the content structure of \( I^{cnt} \) while adhering to the stroke style, texture sparsity, and high-level semantic abstraction of \( I^{ref} \). Additionally, we aim to remove or retain background elements as needed to enhance the foreground object. In this work, we address the sketch generation task using a controllable text-to-image guidance approach. Specifically, we extract object prompts from the content image \( I^{cnt} \) using BLIP~\cite{li2023blip} and incorporate the stroke style and high-level abstraction from the reference sketch \( I^{ref} \) to re-render the final target sketch \( I^{ske} \). To achieve this, we leverage a pre-trained text-to-image model. The overall network architecture is illustrated in Fig~\ref{fig3}, with detailed explanations of each module provided below.

%-------------------------------------------------------------------------
\subsection{Preliminaries}
DDPM inversion~\cite{huberman2024edit} is a process in diffusion-based generative models that reverses denoising steps, enabling the reconstruction of latent representations from generated outputs. Compared to DDIM inversion method~\cite{yang2024mixsa}, DDPM inversion offers greater flexibility for editing tasks by producing noise maps that, while correlated across timesteps and not normally distributed, allow precise image reconstruction and meaningful edits such as color adjustments and structural shifts. A key advantage of DDPM inversion is its ability to maintain an image's structure while altering the conditioning input, such as a text prompt, to enable artifact-free edits that adapt semantics while preserving original details. This efficient method bypasses optimization processes and can enhance diffusion-based editing techniques by improving image fidelity and supporting diverse outputs.

Stable Diffusion~\cite{rombach2022high} incorporates this inversion process within a latent space, rather than a pixel space, increasing computational efficiency and expressiveness. The input image is encoded through a pretrained variational autoencoder (VAE) to produce a latent code \( z \). Denoising then occurs within this latent space using a U-Net architecture that incorporates self-attention and cross-attention mechanisms. Self-attention blocks enhance image detail by calculating attention scores between projected query \( Q \), key \( K \), and value \( V \) vectors:
\begin{align}
A = \text{softmax}\left(\frac{Q \cdot K^\top}{\sqrt{d}}\right), \quad \phi = A \cdot V,
\label{eq1}
\end{align}
where \( A \) is the attention map, \( d \) is the dimensionality, and \( \phi \) is the output of the self-attention block. Cross-attention blocks incorporate conditioning inputs (e.g., text prompts) to guide the generation process.

Finally, the refined latent representation \( z \) is decoded back into an RGB image using the VAE decoder, yielding a high-quality output guided by the conditioning input.

%-------------------------------------------------------------------------
\subsection{Cross-image Stroke Attention}
\label{sec:CSAM}

As discussed in the Introduction, sketch generation as a specialized form of style transfer requires attention to local stroke attributes and consistent texture abstraction across different levels of semantic sparsity. Edge detection serves as the foundation for this task, with edges defining content contours as the most effective strategy for preserving structural information. We extract the contour \( I^{cont} \) from the content image using TEED~\cite{soria2023tiny} and obtain the inverted latents \( z^{cnt}_{T} \), \( z^{ref}_{T} \), and \( z^{cont}_{T} \) for the content image, reference sketch, and contour image using DDPM inversion. During the denoising process, the latent \( z^{cnt}_{T} \) from the content image serves as the initial noise \( z^{ske}_{T} \) for sketch denoising. At each timestep \( t \), we apply Equation~\ref{eq1}.

To achieve effective stroke feature transfer, we utilize latent representations for the content, reference, and contour images obtained through DDPM inversion. For predefined timesteps \( t \in \{0, \ldots, T\} \), the reference sketch \( I^{ref} \), content image \( I^{cnt} \), and contour image \( I^{cont} \) are inverted from their initial state (\( t = 0 \)) to Gaussian noise (\( t = T \)). During DDPM inversion, we also collect the query features of the content (\( Q^{cnt}_t \)), and the key and value features of the reference sketch (\( K^{ref}_t \), \( V^{ref}_t \)) at each timestep.

We initialize the latent noise \( z^{ske}_{T} \) for the stylized sketch by copying the content latent noise \( z^{cnt}_{T} \). To blend features from all three images, we combine the reference keys and values \( K^{ref}_t \) and \( V^{ref}_t \) with the content features. This integration is achieved by mixing the keys and values using the following formulation:
\begin{align}
K^{\text{ske}}_t &= K^{\text{ref}}_t + \alpha K^{\text{cnt}}_t, \\
V^{\text{ske}}_t &= V^{\text{ref}}_t + \alpha V^{\text{cnt}}_t,
\end{align}
where \(\alpha\) is a scalar parameter that control the mixing ratio of content and reference features, allowing flexible adaptation to different stroke characteristics in the sketch.

\begin{figure}[t]
  \includegraphics[width=\linewidth]{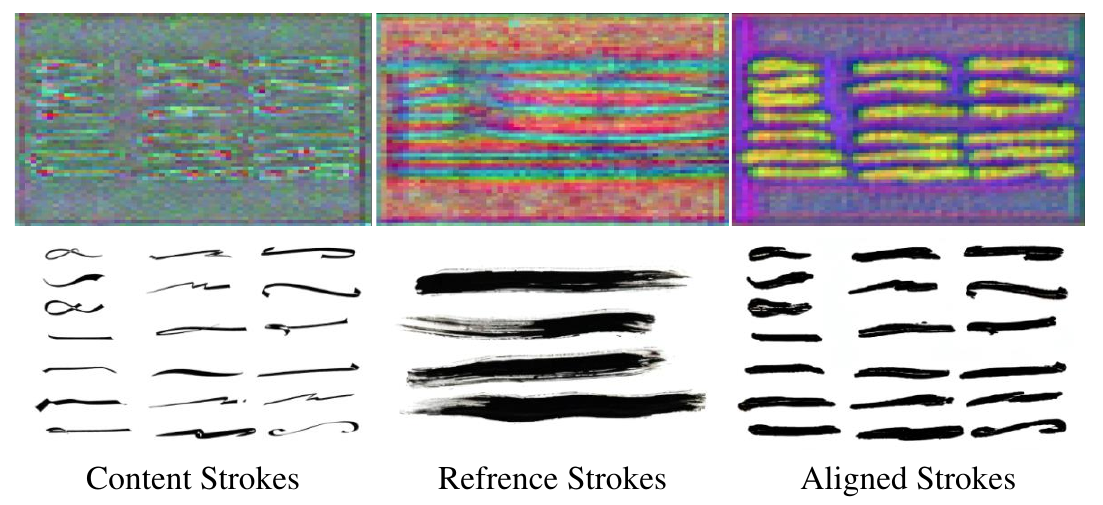}
  \vspace{-5mm}\caption{Stroke alignment results with CSA. The feature images are shown in the first row. The second row shows results where feature alignment between the key and value exchange is applied.}
  \label{fig4}
  \vspace{-4mm}
\end{figure}

 By exchanging and blending features from the content, reference, and contour images, this approach enables the effective transfer of stroke attributes and semantic alignment. However, we found that some strokes may fail to represent the intended content curves accurately (see Fig.~\ref{fig4}). To improve output quality and highlight the foreground, we introduce additional mechanisms to guide sketch generation, as detailed below.

%-------------------------------------------------------------------------
\subsection{Directive Attention Module}
\label{sec:DAM}

\begin{figure}[t]
  \centering
  \includegraphics[width=\linewidth]{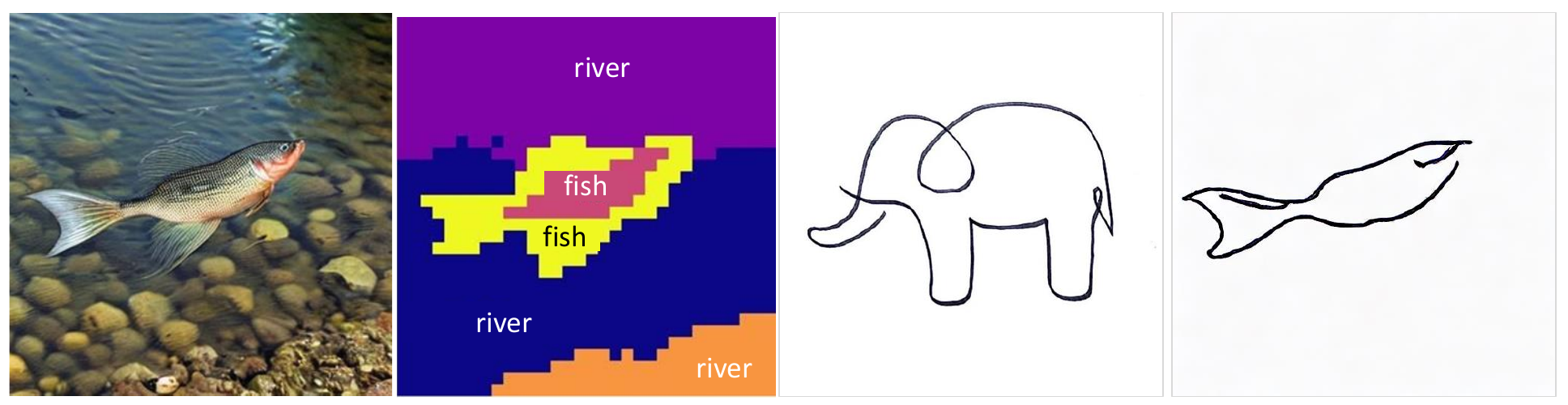}
  \caption{Examples of DAM in action. The first column shows the content images, the second column displays segmentation maps obtained by clustering self-attention maps, the third column provides the reference sketches, and the fourth column illustrates the sketches generated by DAM after applying stroke attribute transfer and segmentation.}
  \label{fig5}
  \vspace{-0.3cm}
\end{figure}

To concentrate stylization on perceptually significant regions, the DAM enhances foreground focus during stroke transfer. We extract self-attention feature maps \( F_{SA} \) at 32×32 resolution, aggregating them across channels via averaging. These maps are segmented into clusters \( M_j \) using KMeans clustering. For each cluster \( j \) and noun \( n \) extracted from the content image’s caption (via BLIP~\cite{li2023blip}), we calculate a relevance score:
\begin{align}
r(j,n) = \frac{\sum_{(x,y)} M_j(x,y) \cdot A_n(x,y)}{\sum_{(x,y)} M_j(x,y) + \delta},
\end{align}
where \( A_n \) is the cross-attention map for noun \( n \), and \( \delta = 10^{-5} \) stabilizes the computation. Clusters with \( r(j,n) > 0.35 \) are designated as foreground regions; remaining areas are suppressed, directing stroke stylization to salient elements (Fig.~\ref{fig5}).

By integrating the segmented self-attention mechanism with cross-image attention, DAM allows precise control over foreground regions, ensuring that stylistic features, such as line styles and high-level semantic abstractions, are faithfully transferred from the reference sketch to the content image. This approach achieves high fidelity in style and content alignment, yielding sketches that closely reflect the intended reference style with minimized background interference. 

\begin{figure}[t]
  \includegraphics[width=\linewidth]{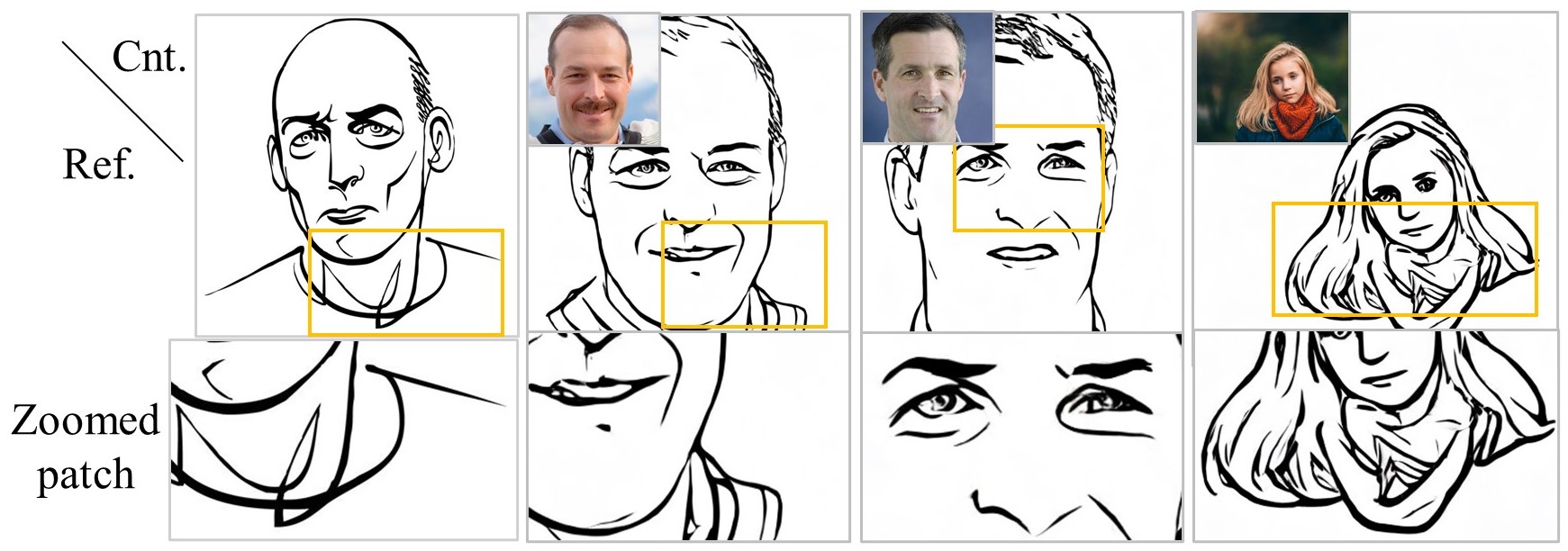}
  \vspace{-5mm}\caption{Comparison of sketches generated using DAM. The top row shows Cnt. and Ref. images with stroke styles reflecting specific high-level attributes such as hair, eyebrows, and abstracted clothing texture. The second to fourth columns display sketches generated to match the content images in the top row, each adopting the stroke styles and high-level features from the reference. The bottom row highlights zoomed-in areas to emphasize the transferred stroke attributes.}
  \label{fig6}
  \vspace{-5mm}
\end{figure}

As illustrated in Fig.~\ref{fig6}, our method successfully aligns high-level semantic features and detailed stroke attributes, producing coherent sketches that maintain stylistic and structural consistency with the reference. This approach minimizes background interference while ensuring that the final sketch reflects the nuanced stroke characteristics of the reference.

%-------------------------------------------------------------------------
\subsection{Semantic Preservation Module}
\label{sec:SPM}

Although the injection of keys and values during sketch generation can effectively transfer stroke attributes to \( I^{ske} \), semantic inconsistencies may arise, particularly when the reference sketch \( I^{ref} \) does not align semantically with the content image \( I^{cnt} \). This can lead to noise and misplaced strokes that disrupt the semantic structure of the generated sketch. To address this, semantic guidance is essential to ensure that each pixel aligns correctly with its corresponding structure.

Text-based guidance alone is often insufficient for precise structural alignment, as textual descriptions may not accurately map to image details. To overcome this limitation, we incorporate contour information to guide semantic pixel alignment and supplement missing semantic strokes. However, we found that overly detailed contour information can disrupt high-level semantics (as edge detection identifies changes based on pixel gradients). For example, when drawing portraits, the eyes are often represented as solid dots rather than pixel-defined circles, necessitating text-based semantic guidance for accurate sketch rendering.

The text-based semantic loss is derived from the guidance provided in the image-to-image diffusion pipeline, which ensures high-level semantic alignment during sketch generation: $L_{sem} = \lambda \cdot \text{CLIP}(I^{ske}, T^{cnt})$, where \( T^{cnt} \) is the text prompt extracted from \( I^{cnt} \) and \( \lambda \) is a weighting parameter for the text guidance.

\begin{figure}[t]
  \includegraphics[width=\linewidth]{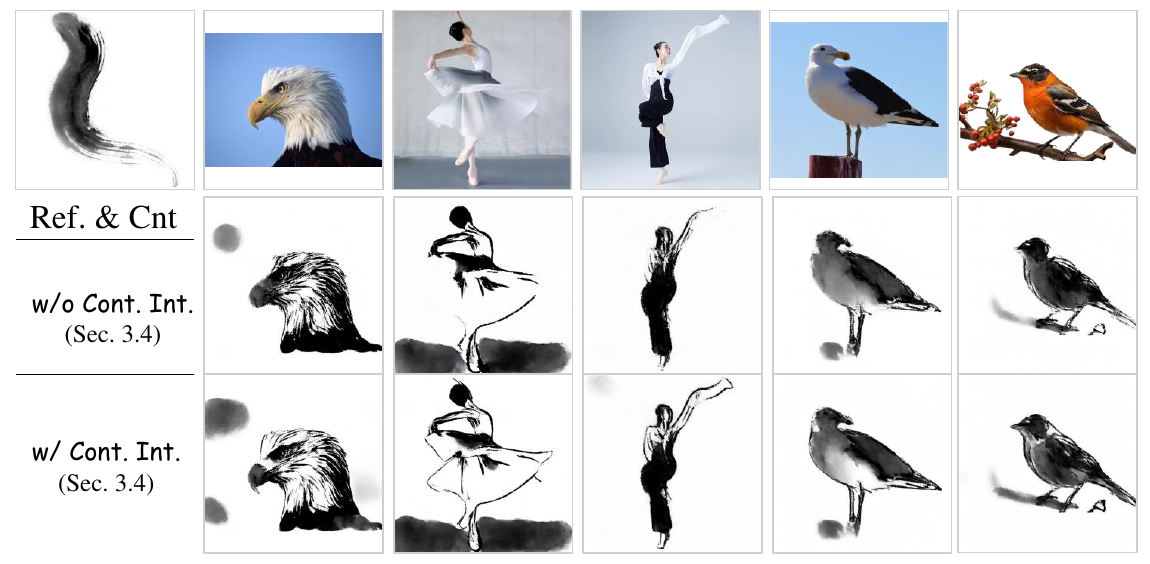}
  \vspace{-5mm}\caption{Qualitative comparison of sketch generation results with and without contour integration. The top row shows Ref. and Cnt. images, followed by results without contour integration (w/o Cont. Int.) and with contour integration (w/ Cont. Int.) as discussed in Sec. 3.4. Contour integration leads to better alignment of semantic features and reduces background interference.}
  \label{fig7}
  \vspace{-0.5cm}
\end{figure}

The contour-based guidance originates from the cached query features during the DDPM inversion process. To integrate this contour information effectively, we use the following equation to adjust the query:
\begin{equation}
Q^{ske}_{i+1} = \gamma  Q^{cont}_{i} + (1 - \gamma) Q^{ske}_{i},
\end{equation}
where \( \gamma \) is a tunable parameter that controls the influence of contour information, set to 0.25 by default in our experiments. As illustrated in Fig.~\ref{fig7}, contour integration significantly improves the alignment of semantic structures, ensuring that key details, such as object outlines, are preserved without introducing unnecessary background noise.

By combining high-level semantic text guidance with contour-based structural alignment, we ensure that the final generated sketch \( I^{ske} \) maintains semantic completeness and accurately represents the image of the content \( I^{cnt} \). This collaborative approach allows for the preservation of both stroke attributes and semantic integrity in the sketch.

%-------------------------------------------------------------------------
\subsection{Stroke Details Propagation Enhancement}
\label{sec:SDPE}

Traditional self-attention blocks often focus on limited areas around image patches, while masked extended attention blocks distribute attention more uniformly by expanding receptive fields across the image. Although this broader approach captures larger context, it can introduce noise and blur finer details \cite{cia, orzech2024masked}. To address this, we adopt a refined contrast operation, inspired by \cite{cia}, to dynamically enhance high-variance regions and suppress low-contrast noise. This contrast adjustment is defined as:
\begin{equation}
\text{Enhance}(A) = (A - \mu(A)) \zeta (\sigma(A)) + \mu(A),
\end{equation}
where \( \mu(A) \) and \( \sigma(A) \) represent the mean and standard deviation operations, respectively, and \( \zeta \) is an adaptive contrast operator. This technique effectively reduces noise, ensuring critical details are preserved during sketch generation.

Building on this, we incorporate the stroke-based refinement concept from SDEdit \cite{meng2022sdedit}. Starting with a sketch initialized by stroke exchange, we denoise the image using classifier-free guidance (CFG) \cite{ho2022classifier, chung2024cfgmc}. During each denoising step, we utilize two parallel forward passes in the network. The first pass employs a cross-image attention layer to capture the stroke characteristics and abstraction level of the reference sketch, generating \( z^{ske} \):$
\epsilon^{\times}_{stroke} = \epsilon^{\times}_{\theta}(z^{ske}_t)$,
while simultaneously retaining the semantic context extracted from the content image’s descriptive prompt:
$\epsilon^{\times}_{text} = \epsilon^{\times}_{\theta}(z^{text}_t)$.
The second pass applies regular self-attention to enhance the sketch’s structural integrity: $\epsilon^{self} = \epsilon^{self}_{\theta}(z^{ske}_t)$.

Following the CFG scheme, the predicted noise \( \epsilon^{t} \) is computed as:
\begin{equation}
\epsilon^{t} = \epsilon^{self} + \beta_{sg}(\epsilon^{\times}_{stroke} - \epsilon^{self}) + \beta_{text}(\epsilon^{\times}_{text} - \epsilon^{self}),
\end{equation}

where \( \beta_{sg} \) is the stroke guidance scale, and \( \beta_{text} \) is the weight for the content text context. %This approach allows us to finely control the alignment of stroke attributes and abstract semantic details from the reference, resulting in stylized sketches that faithfully reproduce both content and style.

%% file: sec/4_experiment.tex
\begin{figure*}[h]
  \includegraphics[width=\linewidth]{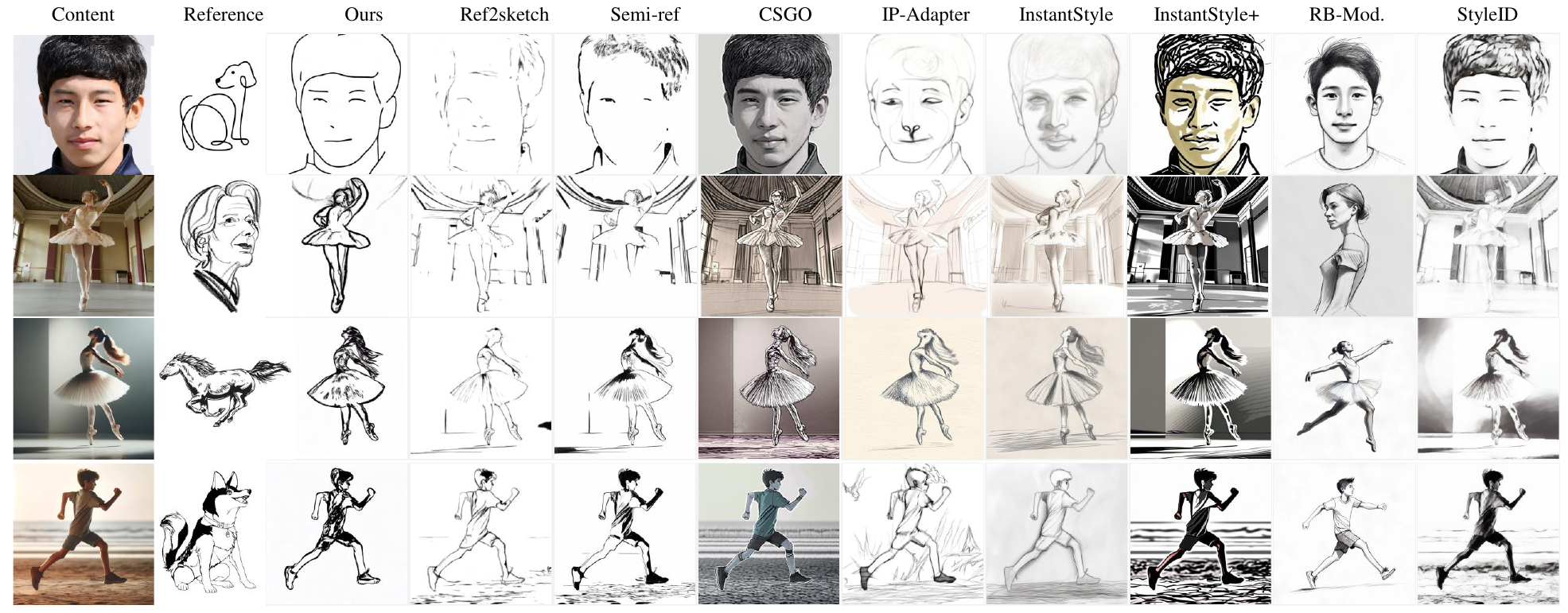}
  \vspace{-5mm}\caption{Qualitative comparison of sketch generation across training-based (4th-6th columns) and training-free baselines (7th-11th columns) using various reference sketches. Our method (3rd column) demonstrates superior adaptability to different reference styles, maintaining both stroke fidelity and semantic consistency across a range of content and reference types.}
  \label{fig8}
    \vspace{-3mm}
\end{figure*}
\section{Experimental Results}
\label{sec:4}
%We conduct all experiments in Stable Diffusion v2-1-base pre-trained model and sampling with $512 \times 512$. For default settings for hyperparameters, we use $\lambda = 0.5$ and $\gamma =0.25$, if they are not mentioned separately.
\textbf{Datasets}. Our experiments utilize three datasets: FS2K~\cite{fan2022facial}, the Anime dataset~\cite{taebum2018anime}, and our newly created Stroke2Sketch-dataset. The FS2K dataset includes 5,140 facial image-sketch pairs. We randomly selected 1,000 pairs for testing, using color images as content images and choosing one sketch from the test set as a reference style. A similar selection strategy was used for the Anime dataset~\cite{taebum2018anime}.

The Stroke2Sketch-dataset includes 50 content images from diverse categories and 20 distinct sketch styles, including single-line sketches, ink sketches, line art, and realistic sketches. Further details are provided in the Appendix.

\noindent\textbf{Metrics}. While traditional metrics like ArtFID~\cite{wright2022artfid}, LPIPS~\cite{zhang2018unreasonable}, and FID~\cite{heusel2017gans} are standard for style transfer, they struggle to capture the semantic sparsity and high-level abstraction unique to sketch generation. Consistent with prior work~\cite{vinker2022clipasso}, we prioritize user perception to better reflect artistic and semantic quality in sketches.

%-------------------------------------------------------------------------

\subsection{Qualitative Comparison}
\label{sec:4.1}

We evaluate our proposed method through a comparison with eight state-of-the-art methods, including three training-based style transfer methods (Ref2sketch~\cite{ref2sketch}, Semi-ref2sketch~\cite{semi2sketch}, and CSGO~\cite{xing2024csgo}), and five training-free style transfer methods (IP-Adapter~\cite{ye2023ip}, InstantStyle~\cite{wang2024instantstyle}, InstantStyle-plus~\cite{insp}, RB-Modulation~\cite{rout2024rbmodulation}, and StyleID~\cite{chung2024style}). Each of these methods takes a reference sketch as input.

\cref{fig8} illustrates the qualitative results across a variety of content images and reference sketches. In the first column, we display the content images, and the second column shows the reference sketches. Notably, the fourth row's reference sketch is of a type seen in the Semi-ref2sketch training data, while the other three reference sketches represent styles outside the training data for Semi-ref2sketch. For methods requiring prompt input, we uniformly set the prompt to ``sketch photo," while other parameters follow each method's default configuration. Our approach demonstrates robustness across diverse reference styles, accurately preserving both the stroke style and high-level semantic details. 

\begin{figure}[t]
  \includegraphics[width=\linewidth]{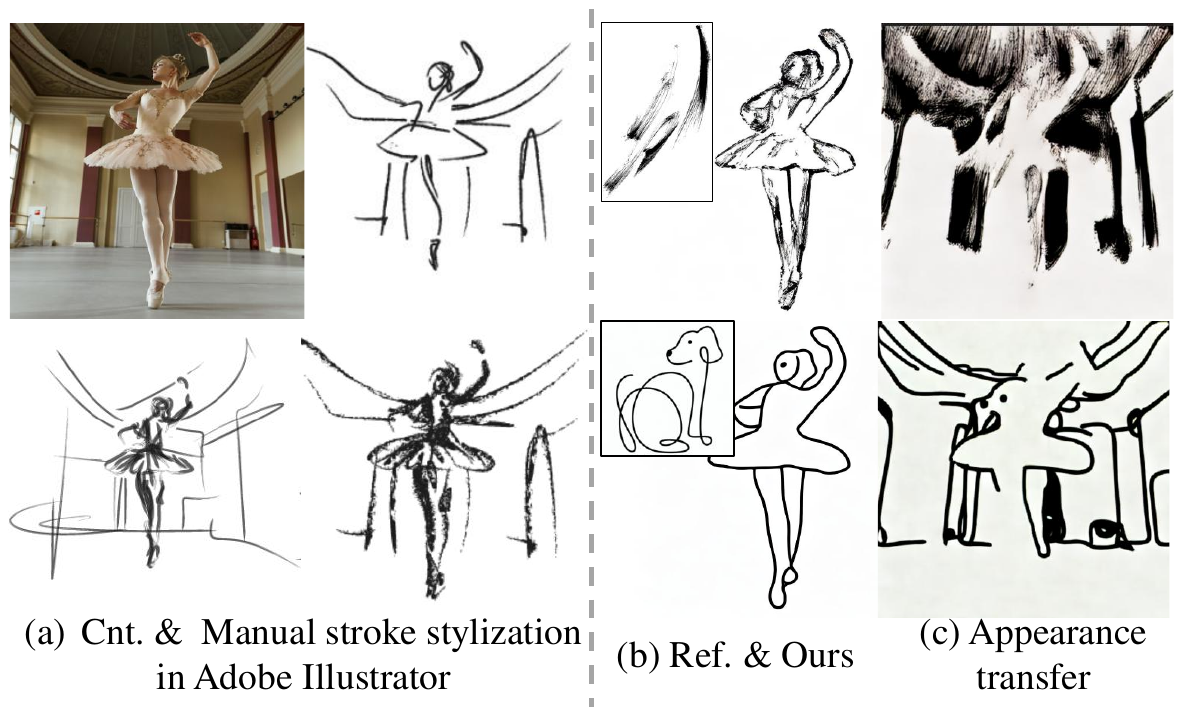}
  \vspace{-5mm}\caption{ (a) Content image and manual stroke stylization by CLIPascene~\cite{vinker2023clipascene} with Adobe Illustrator; (b) Our method’s automatically generated sketches based on reference stroke styles; (c) Appearance transfer results using similar references.}
  \label{fig9}
    \vspace{-4mm}
\end{figure}

\begin{table*}[h]
\centering
\small
\begin{tabular}{l|cccccccc}
\hline
Metric &Ours & Ref2sketch &Semi-ref &Infor-drawing& IP-Adapter &  InstantStyle &InstantStyle+ & StyleID \\ 
\hline
ArtFID $\downarrow$ &\textbf{32.455}  & 45.292 &33.242 &34.214  &33.457&32.532&37.656&35.727\\
LPIPS $\downarrow$ &0.5315 & 0.6982& \textbf{0.5306} & 0.6037&0.6634&0.5432&0.6532&0.5426\\
FID $\downarrow$ &\textbf{22.435}  &34.650   &24.359   &25.035  &24.068&23.940&26.632&25.658 \\
\hline
\end{tabular}
\vspace{-0.1cm}\caption{Quantitative comparison on Stroke2Sketch-dataset with training-based (3rd-5th columns) and training-free baselines (6th-9th columns).}
\label{tab1}
\vspace{-4mm}
\end{table*}

\noindent\textbf{Comparison with vector sketch generation}: \cref{fig9}(a) compares our method with the vectorized sketch generation approach of CLIPascene~\cite{vinker2023clipascene}. The content image is shown alongside vectorized sketches produced by CLIPascene, which were further manually refined with brush strokes in Adobe Illustrator. While CLIPascene can generate abstract sketches, it requires post-processing to achieve consistent stroke styling, whereas our method (\cref{fig9}(b)) automatically produces sketches with stylistically aligned strokes based on reference attributes.

\noindent\textbf{Comparison with appearance transfer methods}: \cref{fig9}(c) shows results from appearance transfer method of CIA~\cite{cia} applied to the same reference sketches. Although CIA excel in transferring visual features based on semantic similarity and object category, they focus on appearance rather than stroke style, making them less suitable for our sketch generation goals.%Additional qualitative comparisons are provided in Appendix C.

\begin{figure}[t]
  \centering
  \includegraphics[width=1\linewidth]{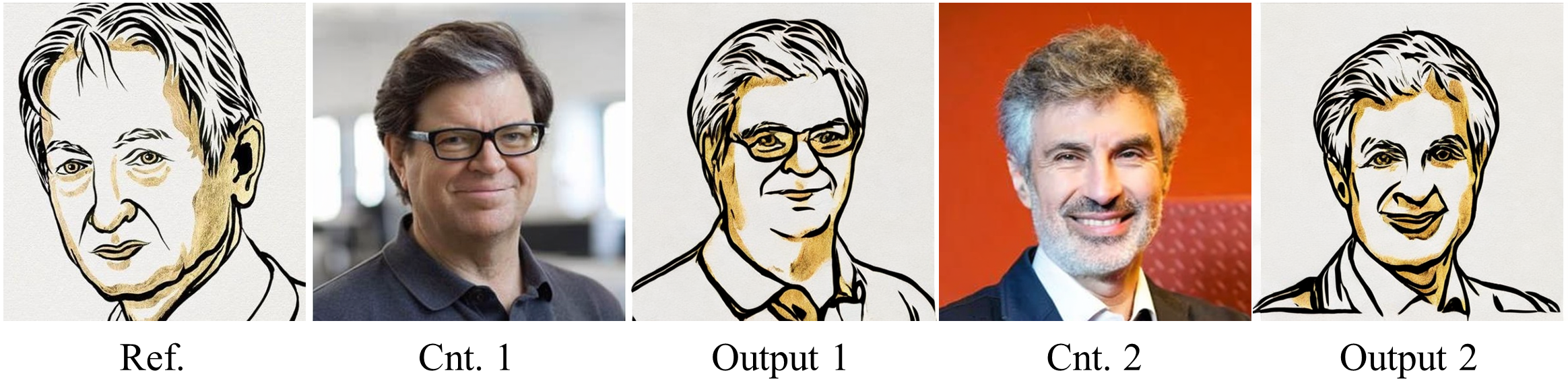}
  \vspace{-4mm}
  \caption{Stroke2Sketch's color sketch generation preserving reference styles and stroke characteristics} 
  \label{fig12_}
  \vspace{-4mm}
\end{figure}

\noindent\textbf{Non-grayscale Sketch Generation.}
Building on its grayscale performance, 
Stroke2Sketch maintains reference stroke patterns and artistic styles in color outputs (Fig.~\ref{fig12_}).
\vspace{-2mm}

%-------------------------------------------------------------------------
\subsection{Quantitative Comparison}
\label{sec:4.2}

We quantitatively evaluate our method against several state-of-the-art sketch extraction techniques, including both training-based (Ref2Sketch~\cite{ref2sketch}, Semi-Ref2Sketch~\cite{semi2sketch}, and Informative-drawing~\cite{chan2022learning}) and training-free style transfer methods (IP-Adapter~\cite{ye2023ip}, InstantStyle~\cite{wang2024instantstyle}, InstantStyle-plus~\cite{insp}, StyleID~\cite{chung2024style}). Table~\ref{tab1} presents the results on the Stroke2Sketch-dataset, showing that our method achieves the lowest ArtFID and FID scores, indicating superior performance in both style alignment and content preservation. These metrics highlight our method's effectiveness in producing high-quality sketches with strong semantic and stylistic fidelity. %Results for the other datasets can be found in Appendix B.

%-------------------------------------------------------------------------
\subsection{User Study}
\label{sec:4.3}
\begin{figure}[t]
  \includegraphics[width=\linewidth]{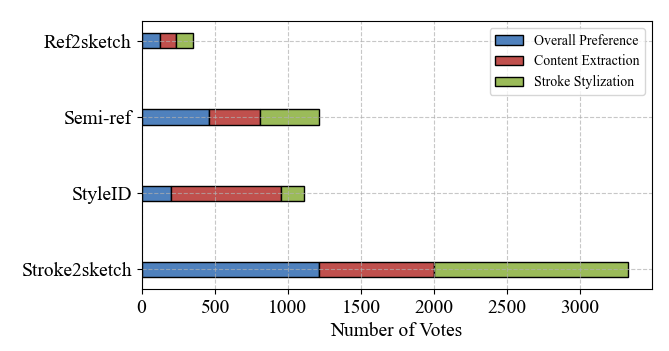}
  \vspace{-5mm}\caption{User study results. Our proposed method received the highest preference. }
  \label{fig10}
    \vspace{-4mm}
\end{figure}

We randomly selected 15 content images and 15 reference sketches from Stroke2Sketch-dataset, creating 225 image pairs. From these, we sampled 20 pairs to generate stylized images using four methods. The results were displayed side-by-side in random order, and participants were asked to choose their favorite based on three criteria: content extraction, stroke stylization, and overall preference. We collected 2,000 votes for each criterion from 100 users and presented the results as a bar chart. As shown in \cref{fig10}, our method received the highest preference, outperforming both training-based and training-free baselines, demonstrating its effectiveness in handling diverse stroke styles and abstract artistic effects.% Further details of the perceptual study are in Appendix B.
 
%-------------------------------------------------------------------------
\subsection{Ablation Study}
\label{sec4.4}
\begin{figure}[t]
  \includegraphics[width=\linewidth]{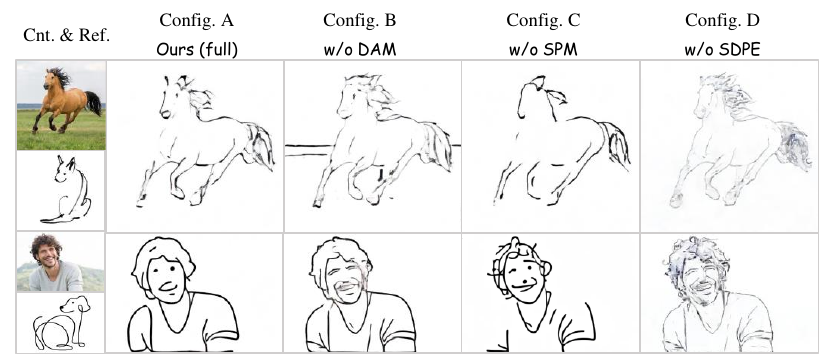}
  \vspace{-5mm}
  \caption{Ablative qualitative comparison of different variants of our method.}
  \label{fig11}
  \vspace{-3mm}
\end{figure}

%To validate our method’s components, we performed an ablation study on the ref2sketch-dataset. As shown in Table~\ref{tab2} and \cref{fig11}, removing any of the Directive Attention Module (DAM), Semantic Preservation Module (SPM), or Stroke Details Propagation Enhancement (SDPE) led to decreased ArtFID, FID, and LPIPS scores, confirming the contribution of each component. All parameter settings and ablation details are in Appendix A.
To validate our method’s components, we performed an ablation study on the Stroke2Sketch-dataset. As shown in Table~\ref{tab2}, removing any of the Directive Attention Module (DAM), Semantic Preservation Module (SPM), or Stroke Details Propagation Enhancement (SDPE) led to decreased performance across ArtFID, FID, and LPIPS metrics, confirming the contribution of each component. Specifically, without DAM, ArtFID increased from 32.45 to 38.67 and FID rose to 26.53, indicating weaker style-content alignment and content leakage in sketches, as observed in \cref{fig11}. The absence of SPM resulted in ArtFID rising to 36.89 and FID to 30.47, with sketches losing semantic coherence and structural integrity, such as poorly defined object outlines. Most notably, removing SDPE caused the sharpest decline, with ArtFID reaching 40.53 and sketches exhibiting excessive noise and loss of fine details, as evidenced by cluttered textures in \cref{fig11}. In contrast, the full method (Configuration A) achieved optimal scores of 32.45 (ArtFID), 22.43 (FID), and 0.530 (LPIPS), producing high-quality, reference-aligned sketches with precise details. These results highlight the critical roles of DAM, SPM, and SDPE in enhancing sketch quality. %For a detailed analysis of each configuration and additional visual comparisons, please refer to Appendix A.
 
\begin{table}[t]
\centering
\small
\begin{tabular}{l|ccc}
\hline
Configuration & ArtFID & FID & LPIPS \\
\hline
A: Ours %($\gamma = 0.25$, default) 
& \textbf{32.45} &\textbf{22.43}&\textbf{0.530} \\
B: - DAM (Sec.~\ref{sec:DAM}) &38.67 & 26.53 & 0.672 \\
C: - SPM (Sec.~\ref{sec:SPM}) &36.89 & 30.47 & 0.637 \\
D: - SDPE (Sec.~\ref{sec:SDPE}) & 40.53 & 32.44 & 0.598 \\
\hline
\end{tabular}
\caption{Ablation study of different variants of our method.}
\label{tab2}
\vspace{-0.2cm}
\end{table}

%% file: sec/5_discussion.tex
\section{Conclusion}
\label{sec:formatting}

Our training-free method for content-to-sketch generation aligns stroke attributes and semantic texture sparsity, mimicking artistic subject extraction to produce high-fidelity sketches. Leveraging pretrained models, it achieves state-of-the-art performance in quantitative metrics and user evaluations. Limitations arise with overly simplistic or complex reference sketches (see appendix for failure cases). Future work could explore decoupling semantic information from stroke attributes to improve adaptability and sketch quality.

\noindent\textbf{Acknowledgment.} This work is supported by the Guangdong Natural Science Funds for Distinguished Young Scholars (Grant 2023B1515020097), the National Research Foundation, Singapore under its AI Singapore Programme (AISG Award No.: AISG3-GV-2023-011), the Singapore Ministry of Education AcRF Tier 1 Grant (Grant No.: MSS25C004), and the Lee Kong Chian Fellowships.

%% file: sec/suppl.tex
\clearpage
\maketitlesupplementary

%\pagenumbering{roman}
%\setcounter{figure}{0}
\setcounter{section}{0}
\renewcommand{\thesection}{\Alph{section}}
\renewcommand{\thesubsection}{\thesection .\arabic{subsection}} 

%\tableofcontents

\section{Analysis and ablation}\label{sec:a}
\subsection{Stroke stylization}\label{sec:a1}

One of the main challenges in sketch extraction is how to transfer stroke attributes from a reference sketch to reconstruct the content image's sketch. As discussed in the main paper's related work section, previous approaches often rely on algorithmic simulations to emulate specific stroke styles. However, the vast diversity of sketch styles in real-world references makes it impractical to enumerate and simulate all possible styles algorithmically.

\begin{figure}[h]
  \centering
  \includegraphics[width=\linewidth]{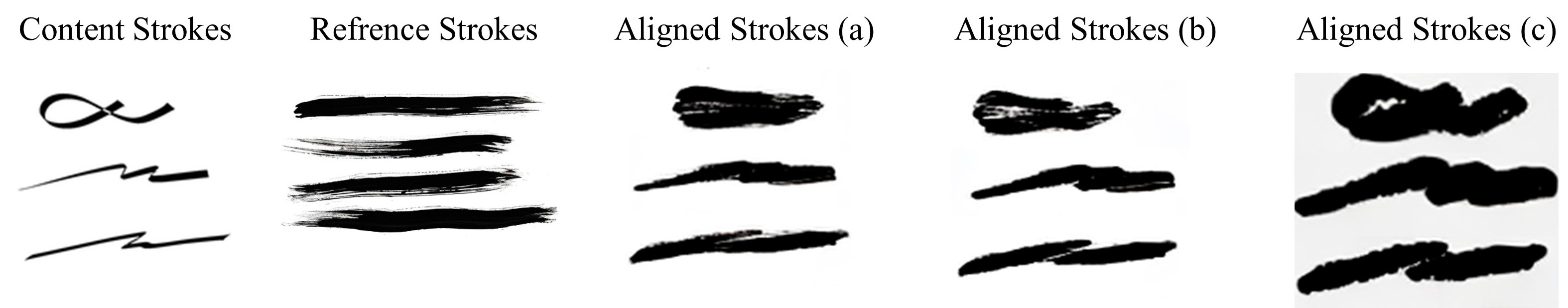}
  \caption{Stroke alignment results. The first two columns show the content strokes and reference strokes, respectively. Column (a) displays results with direct K-V exchanges, showing partial curve distortion. Columns (b) and (c) show improvements using contour guidance and stroke details propagation enhancement, respectively, highlighting the balance between stroke consistency and content preservation.}
  \label{fig12}
\end{figure}

Our proposed approach introduces a novel solution by leveraging key-value (K-V) exchanges in attention mechanisms to transfer stroke attributes. This method allows dynamic adaptation of reference stroke properties to the content sketch during the generation process. However, as shown in the third column of \cref{fig12} (a), direct K-V exchanges can sometimes distort structural elements, such as curves, leading to incomplete or misaligned strokes.

To address these limitations, we integrate contour guidance and the SDPE module into the generation process. These enhancements enable the system to retain structural integrity while achieving stroke style consistency. As demonstrated in \cref{fig12}, column (b) shows results with contour guidance applied, which helps preserve critical outlines while aligning strokes. Column (c) illustrates the output with both contour guidance and SDPE, achieving a balance between stroke stylization and content preservation.

While these methods improve stroke consistency, they can occasionally compromise the semantic expression of the content. To mitigate this, we introduce user-adjustable parameters, allowing users to fine-tune the balance between style fidelity and content preservation based on specific application requirements. In the following section, we detail the default parameters used in our experiments and provide the rationale for their selection.

\subsection{Experimental configuration}\label{sec:a2}

We operate using the Stable Diffusion v2.1-base model\footnote{\url{https://huggingface.co/stabilityai/stable-diffusion-2-1-base}}~\cite{rombach2022high}, leveraging DDPM inversion~\cite{huberman2024edit} for input image inversion and the DDIM scheduler for denoising over 50 steps. Following~\cite{cia}, cross-image attention layers are employed at specific resolutions (32×32 and 64×64) during denoising, enhancing stroke injection.  The injection timesteps and additional settings are summarized in ~\cref{tab:config}. Further, object prompts are extracted using BLIP-2\footnote{\url{https://huggingface.co/docs/transformers/main/model_doc/blip-2}}~\cite{li2023blip}, and contour detection is performed using TEED~\cite{soria2023tiny} and U2-Net\footnote{\url{https://github.com/xuebinqin/U-2-Net}}. To ensure semantic segmentation, the unsupervised self-segmentation technique from~\cite{patashnik2023localizing} is applied. 

\begin{table}[h]
\centering
\begin{tabular}{l|l}
\hline
\textbf{Hyperparameter}         & \textbf{Value/Methodology} \\ \hline
\textbf{Model}                  & Stable Diffusion v2.1-base\footnotemark[1] \\
Inversion                       & DDPM inversion~\cite{huberman2024edit} \\
Denoising Scheduler             & DDIM, 100 steps (30 steps skip) \\
\multirow{2}{*}{Resolution for SFI} & 32×32 (steps 10–70)\\& 64×64 (steps 10–90) \\
Contrast Strength               & $\zeta=1.67$ \\
Contour Mask                    & U2-Net\footnotemark[3]\\    
Contour Detection               & TEED~\cite{soria2023tiny} \\
\multirow{2}{*}{Guidance Scales}& $\beta_{sg}=5$, $\beta_{text}=0.1$\\& (steps 20–100) \\
Self-Segmentation               & Patashnik et al.~\cite{patashnik2023localizing} \\
Contour Guidance                & $\gamma=0.25$ \\
Prompt Extraction               & BLIP-2\footnotemark[2]~\cite{li2023blip} \\
Device                          & CUDA NVIDIA RTX 3090\\
Seed                            & 42 \\
\hline
\end{tabular}
\caption{Hyperparameter settings for Stroke2Sketch experiments.}
\label{tab:config}
\end{table}
\subsection{Ablation study analysis}\label{sec:a3}

As discussed in \cref{sec4.4} of the main paper, we performed ablation studies to validate the contributions of the DAM, SPM, and SDPE. Quantitative results in \cref{tab2} and qualitative comparisons in \cref{fig11} demonstrate the critical roles of these components in achieving high-quality sketch generation.

Removing any component results in significant performance degradation, as reflected in both metrics and visual outputs:

\textbf{Configuration B: Without DAM.} Removing DAM results in ArtFID increasing from 32.45 to 38.67 and FID increasing from 22.43 to 26.53, indicating weaker style-content alignment and semantic consistency. LPIPS worsens to 0.672, highlighting the loss of content fidelity. Visually, as shown in \cref{fig11}, the absence of DAM causes noticeable content leakage, leading to inconsistent stroke thickness and blurred object boundaries. For example, the foreground details, such as facial contours and clothing edges, become misaligned, disrupting the overall semantic clarity.

\textbf{Configuration C: Without SPM.} Without SPM, ArtFID increases to 36.89, FID worsens to 30.47, and LPIPS rises to 0.637, reflecting reduced semantic alignment. \cref{fig11} shows that this configuration struggles to preserve high-level abstractions, with many fine details either omitted or misplaced. For instance, the strokes in object outlines lose coherence, and elements such as eyes or limbs become poorly defined. This highlights the importance of SPM in maintaining semantic coherence and ensuring structural integrity.

\textbf{Configuration D: Without SDPE.} The removal of SDPE leads to the most significant degradation, with ArtFID increasing to 40.53 and FID and LPIPS scores worsening to 32.44 and 0.598, respectively. Visually, \cref{fig11} reveals that sketches become overly coarse and noisy, with significant background interference and a lack of refinement in stroke details. For example, small textures and edges appear cluttered, reducing the clarity and aesthetic quality of the sketch. SDPE is essential for refining fine-grained details and suppressing noise propagation.

\textbf{Configuration A: Full Method.} The full method achieves the best performance, with ArtFID, FID, and LPIPS scores of 32.45, 22.43, and 0.530, respectively. Qualitatively, as seen in \cref{fig11}, this configuration produces sketches that closely align with the reference stroke style while preserving the semantic structure of the content. Fine details, such as facial features and object edges, are rendered with high precision, demonstrating the effectiveness of integrating DAM, SPM, and SDPE.

\begin{figure}[h]
  \centering
  \includegraphics[width=\linewidth]{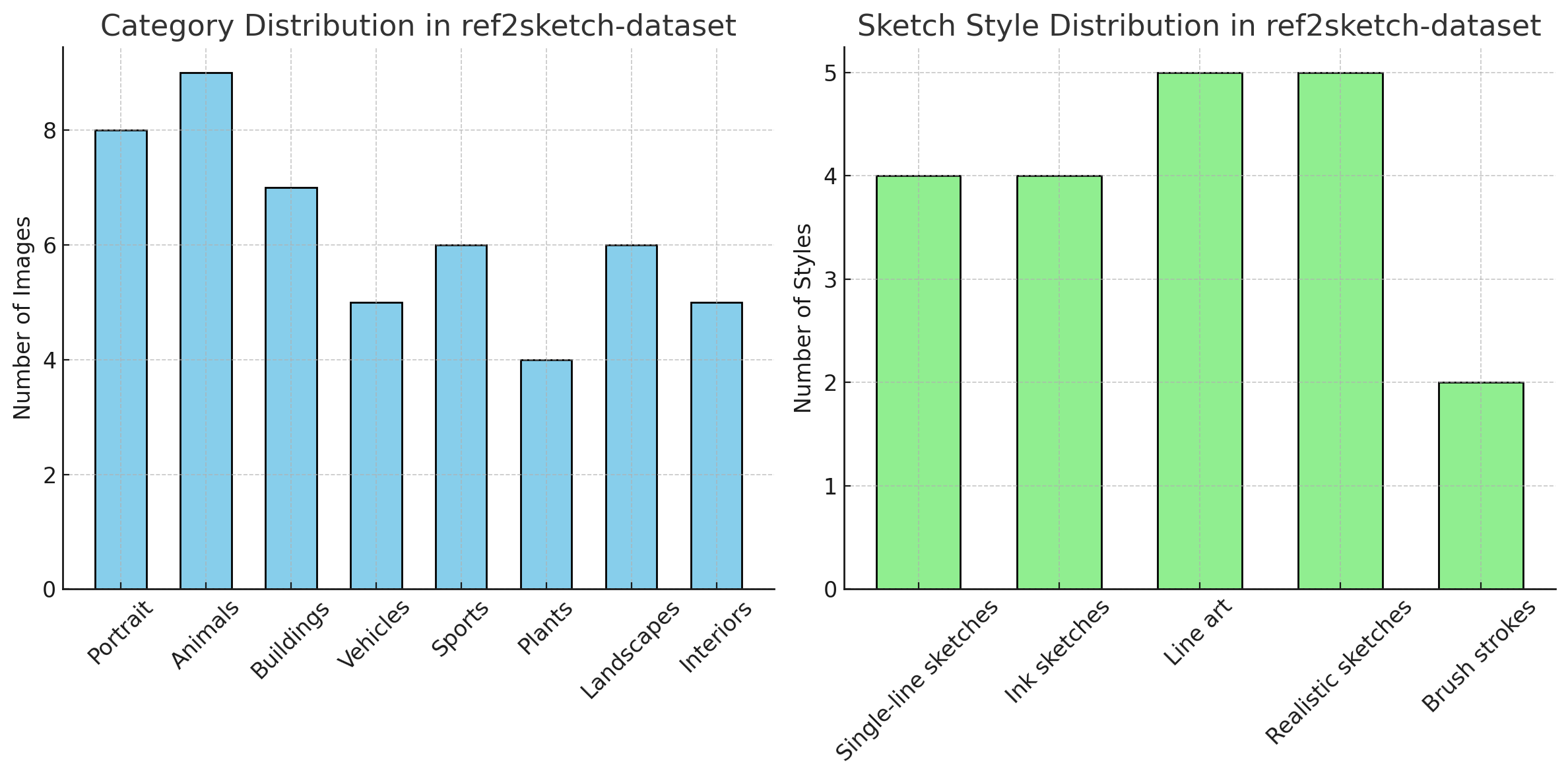}
  \caption{Overview of the Stroke2Sketch-dataset: Left - category distribution; Right - sketch style distribution. Zoom in to view details.}
  \label{fig21}
\end{figure}
\subsection{Hyperparameter effects}

\begin{figure*}[h]
  \centering
  \includegraphics[width=0.9\linewidth]{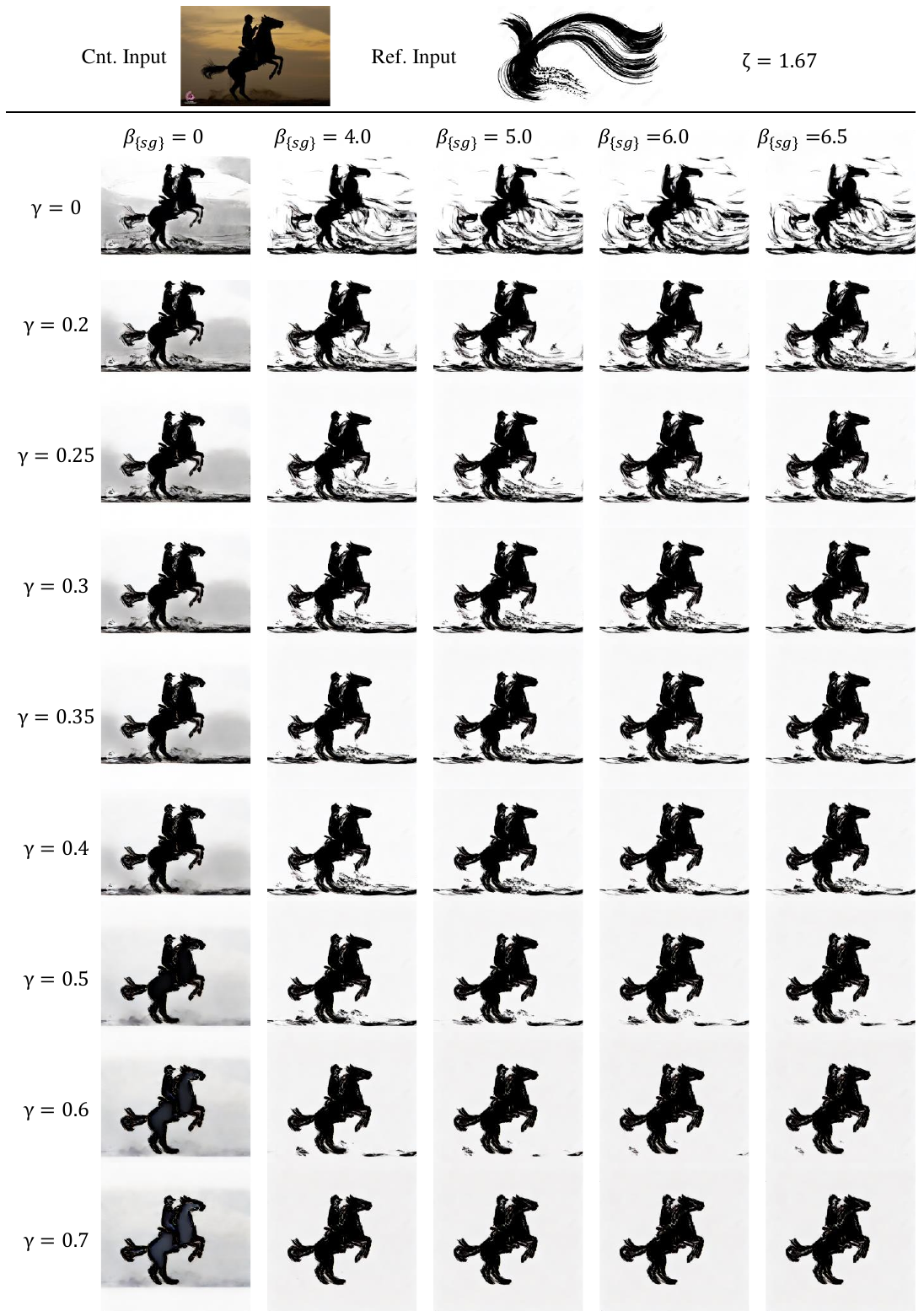}
  \caption{Visualization of \(\gamma\) variations. Increasing \(\gamma\) improves contour alignment but reduces stylistic abstraction. Default setting: \(\gamma = 0.25\). Zoom in to view details.}
  \label{fig13}
\end{figure*}

\begin{figure*}[h]
  \centering
  \includegraphics[width=\linewidth]{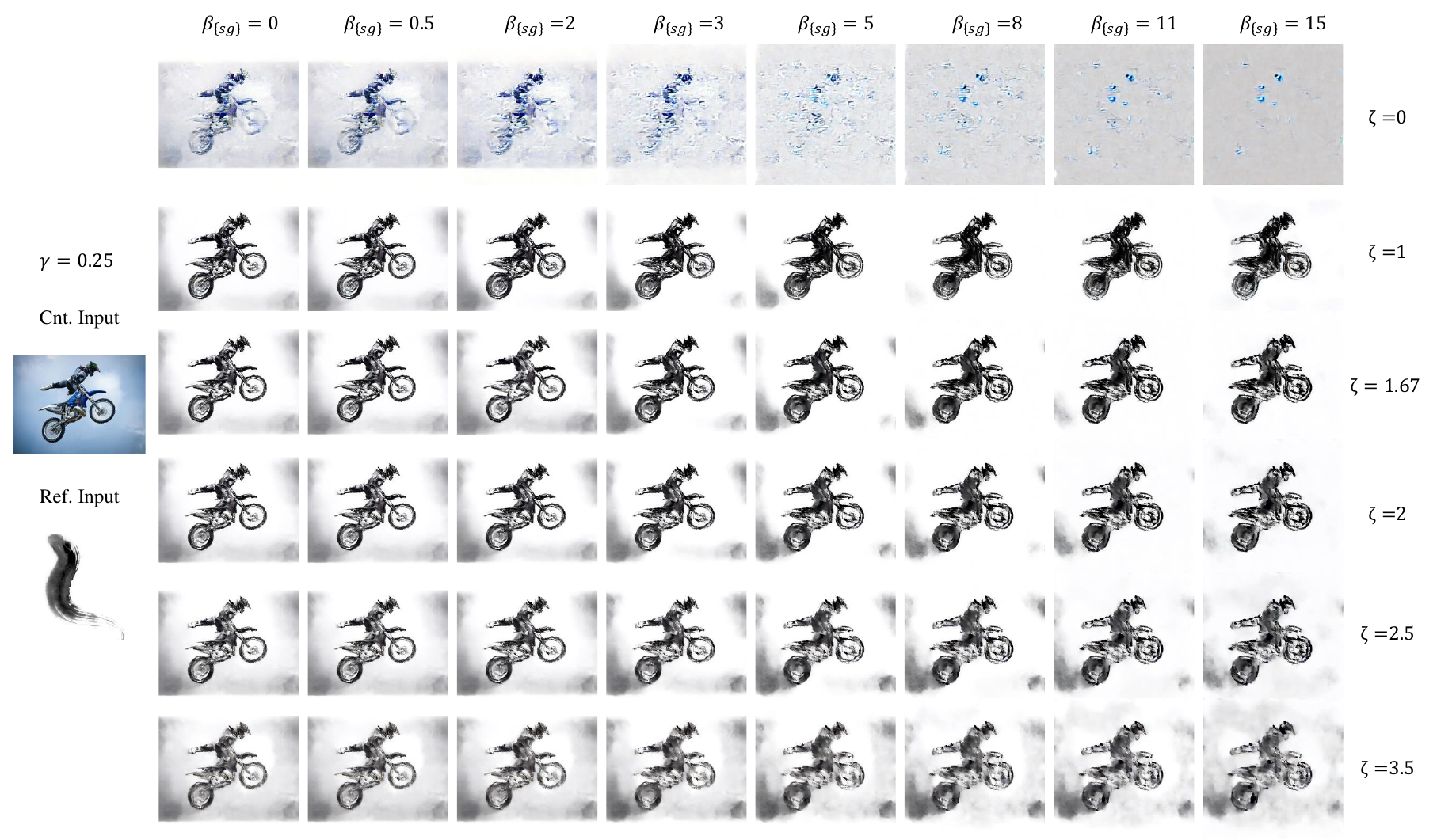}
  \caption{Visualization of \(\beta_{sg}\) variations. Higher \(\beta_{sg}\) emphasizes stroke attributes but may diminish content fidelity. Default setting: \(\beta_{sg} = 5\). Zoom in to view details.}
  \label{fig14}
\end{figure*}

\begin{figure*}[h]
  \centering
  \includegraphics[width=\linewidth]{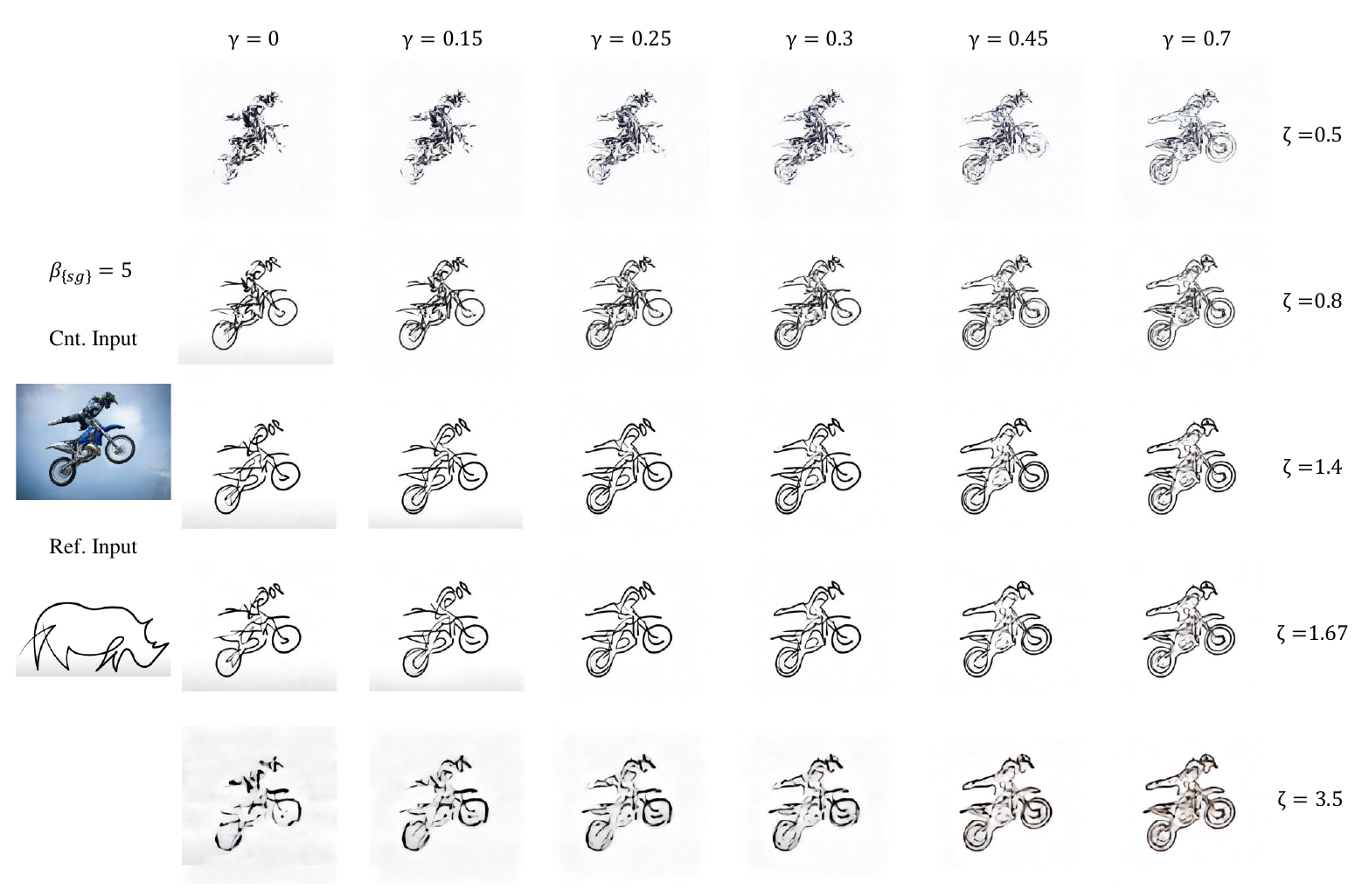}
  \caption{Visualization of \(\zeta\) variations. Optimal contrast strength is achieved at \(\zeta = 1.67\). Excessive \(\zeta\) introduces over-sharpening effects. Zoom in to view details.}
  \label{fig15}
\end{figure*}

We demonstrate in \cref{fig13},~\cref{fig14}, and~\cref{fig15} how varying the hyperparameters \(\gamma\), \(\beta_{sg}\), and \(\zeta\) provides users with greater control over the sketch generation process. These parameters influence the balance between style fidelity, content preservation, and abstraction, enabling customization based on specific user needs. Observing the results across various sketches, we note the interplay of these parameters with the pretrained diffusion model priors and the initial contour extraction quality.

\textbf{Effect of \(\gamma\) (Contour weight):} The parameter \(\gamma\) determines the influence of content image contours on the final sketch. As shown in ~\cref{fig13}, increasing \(\gamma\) results in sketches with more pronounced alignment to the original content structure, improving realism. For example, at \(\gamma = 0.25\) (our default setting), the contours are well-preserved while maintaining the reference stroke style. However, higher values of \(\gamma\) (e.g., \(\gamma = 0.6\)) lead to excessive adherence to the content outline, compromising the transfer of stylistic features. Conversely, very low values (e.g., \(\gamma = 0.15\)) result in sketches with diminished structural coherence, favoring abstraction.

\textbf{Effect of \(\beta_{sg}\) (Stroke guidance scale):} The parameter \(\beta_{sg}\) controls the weight of stroke attributes transferred from the reference image. In ~\cref{fig14}, we observe that lower values of \(\beta_{sg}\) (e.g., \(\beta_{sg} = 2\)) yield sketches with reduced stylization, leaning more toward content fidelity. As \(\beta_{sg}\) increases, the reference stroke features become more prominent, with the optimal balance achieved at \(\beta_{sg} = 5\). However, excessively high values (e.g., \(\beta_{sg} = 15\)) can lead to exaggerated stylization, overshadowing the content image’s structural elements.

\textbf{Effect of \(\zeta\) (Contrast strength):} The parameter \(\zeta\) enhances contrast in the attention maps, aiding in stroke detail refinement. As shown in ~\cref{fig15}, low values of \(\zeta\) (e.g., \(\zeta = 0.8\)) result in sketches with softer, less defined strokes. The default setting (\(\zeta = 1.67\)) provides a balanced output with clear stroke details and stylistic alignment. Increasing \(\zeta\) beyond 3.5 introduces over-sharpening, leading to unnatural and overly rigid strokes.

\textbf{Combined effects and user control:} By varying these parameters in combination, users can control the degree of abstraction and stylization. For instance, increasing \(\gamma\) while decreasing \(\beta_{sg}\) emphasizes content realism, which is suitable for architectural sketches. In contrast, lowering \(\gamma\) and increasing \(\beta_{sg}\) enhances artistic abstraction, ideal for expressive line art. Default settings of \(\zeta = 1.67\), \(\gamma = 0.25\), and \(\beta_{sg} = 5\) provide a general-purpose configuration that balances stroke style consistency with content preservation. Users can further refine these parameters based on their specific objectives.

\section{Evaluation details}\label{sec:b}
\subsection{Stroke2Sketch-dataset}\label{sec:b1}

As described in the main paper, the Stroke2Sketch-dataset was created to assess the human perception of different sketch extraction methods. \cref{fig21} provides a detailed visualization of the category distribution and sketch style diversity in the ref2sketch-dataset. This comprehensive dataset serves as a benchmark for evaluating both stylistic fidelity and semantic alignment in sketch generation tasks.

\subsection{Baseline implementations}\label{sec:b2}
 When comparing to alternative methods, we used the following implementations or demo websites:
 \begin{itemize}
     \item Ref2sketch: \href{https://github.com/ref2sketch/}{https://github.com/ref2sketch/ref2sketch}
     \item Semi-ref2sketch: \href{https://github.com/Chanuku/semi_ref2sketch_code}{https://github.com/Chanuku/semi\_ref2\\sketch\_code}
     \item Informative-drawings: \href{https://github.com/carolineec/informative-drawings}{https://github.com/carolineec/infor\\mative-drawings}
     \item IP-Adapter: \href{https://github.com/tencent-ailab/IP-Adapter}{https://github.com/tencent-ailab/IP-Adapter}
     \item InstantStyle: Huggingface demo \href{https://huggingface.co/spaces/InstantX/InstantStyle}{https://huggingface.co/spaces/InstantX/InstantStyle}
     \item InstantStyle-plus: \href{https://github.com/instantX-research/InstantStyle-Plus}{https://github.com/instantX-research/InstantStyle-Plus}
     \item CSGO: Huggingface demo \href{https://huggingface.co/spaces/xingpng/CSGO}{https://huggingface.co/spaces/xingpng/CSGO}
     \item RB-Modulation: Huggingface demo \href{https://huggingface.co/spaces/fffiloni/RB-Modulation}{https://huggingface.co/spaces/fffiloni/RB-Modulation}
\end{itemize}

\subsection{Quantitative results on Stroke2Sketch-dataset}\label{sec:b3}

As shown in \cref{tab1} in the main paper, our method achieves the lowest ArtFID and FID values among both training-based and training-free baselines, demonstrating its superiority in style fidelity and content preservation. Although our LPIPS value is slightly higher than Semi-ref2sketch~\cite{semi2sketch}, this discrepancy is expected due to the unique emphasis on stroke consistency in our approach. Notably, LPIPS, as a pixel-level similarity metric, does not fully capture the complexity of reference-based sketch extraction, where abstract artistic effects and semantic alignment are crucial. This limitation is evident in user evaluations, where our method consistently outperforms baselines, as detailed in \cref{sec:4.2} of the main paper.

Informative-drawings~\cite{chan2022learning}, designed to work with predefined styles, performs well on similar styles but lacks the flexibility to generalize to arbitrary reference sketches. 

\subsection{Quantitative results on FS2K dataset}\label{sec:b4}

In addition to the Stroke2Sketch-dataset, we evaluated our method on the FS2K dataset. \cref{tab4} highlights our method’s superior performance compared to specialized sketch extraction methods (Ref2sketch~\cite{ref2sketch}, Semi-ref2sketch~\cite{semi2sketch}) and recent style transfer methods (StyleID~\cite{chung2024style}). Our method achieves the lowest FID (128.84) and LPIPS (0.4057) values, showcasing its robustness in producing high-quality sketches with strong semantic and stylistic fidelity. 

While Ref2sketch and Semi-ref2sketch demonstrate reasonable performance due to their focus on training with paired data, they lack the flexibility to adapt to varied and abstract reference sketches. StyleID, although effective in style transfer tasks, struggles with precise alignment when handling content-specific sketches. In contrast, our approach leverages contour guidance and cross-image attention to preserve both structural details and stylistic nuances, ensuring high-quality results even in complex scenarios.
\begin{table}[h]
\centering
\begin{tabular}{l|ll}
\hline
\textbf{Methods}         & LPIPS &FID \\ \hline
Ref2sketch &0.5309&228.15\\ 
Semi-ref2sketch &0.4540&185.26\\ 
StyleID &0.5494&208.64\\
Ours & \textbf{0.4057}&\textbf{128.84}\\
\hline
\end{tabular}
\caption{ Quantitative results of comparison with baselines on FS2K dataset}
\label{tab4}
\end{table}

\subsection{Perceptual Study}\label{sec:b5}
\begin{figure*}
    \centering
    \includegraphics[width=0.95\linewidth]{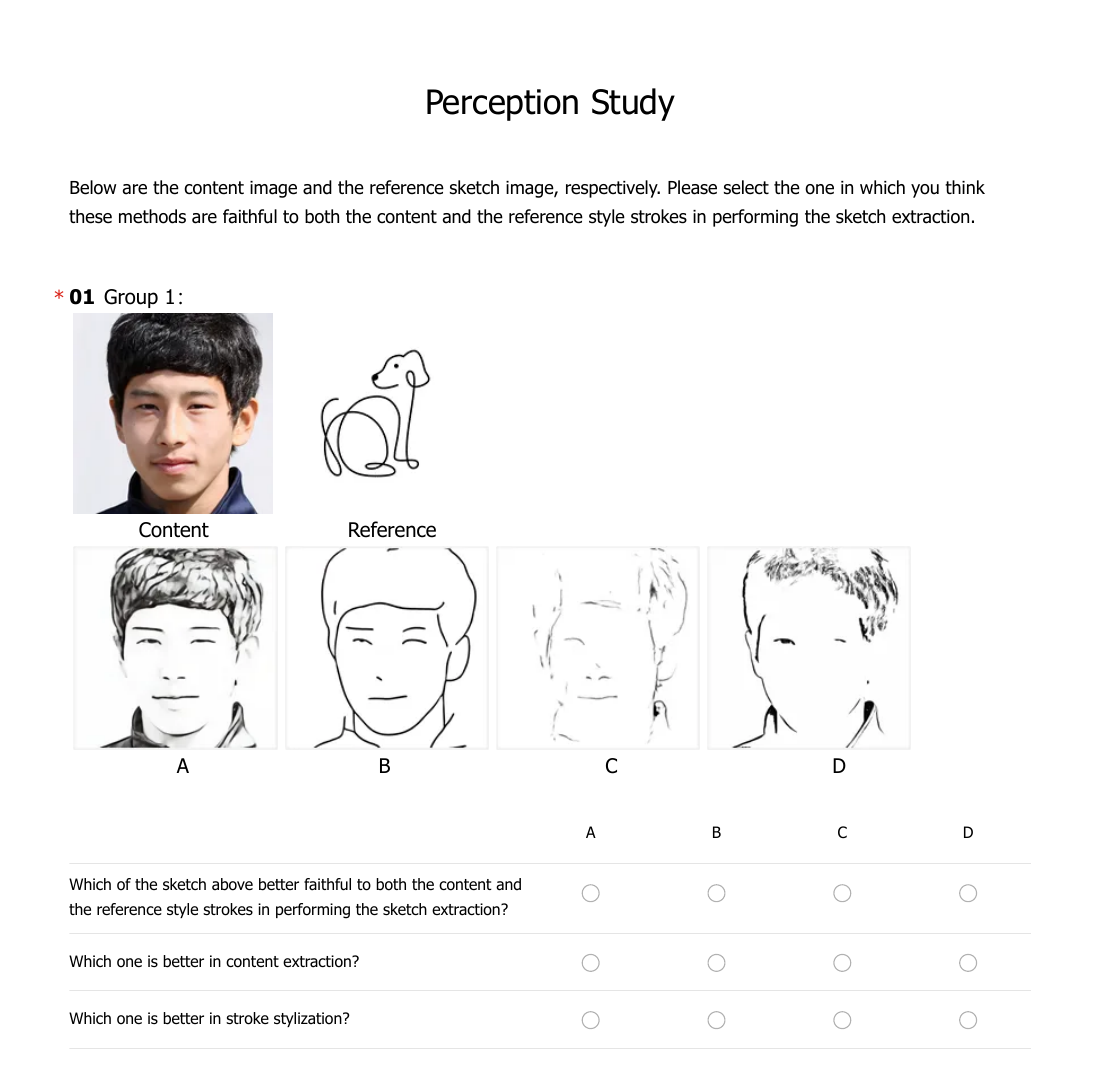}
    \caption{Designed user study interface. }
    \label{fig16}
    %\vspace{-4mm}
\end{figure*}

Our user study interface (\cref{fig16}) displays the source content-reference pair as visual anchors alongside four anonymized stylized results in randomized layouts. Participants independently evaluated 20 unique image pairs, with each session limited to 5 minutes to ensure focused judgments. The interface incorporated a training phase showing prototypical examples of high/low content extraction and stroke quality before formal evaluation. We implemented quality control by tracking response times (excluding votes $< 3s$ as rushed) and adding attention-check questions. Detailed voting distributions per image pair and participant demographic profiles (85\% with art-related backgrounds) are archived in the supplemental material.
\section{Additional Results}\label{sec:c}

As discussed in \cref{sec:4.1} of the main paper, we compare Stroke2Sketch with eight state-of-the-art methods that support both reference-based and text-based inputs, ensuring a fair evaluation of our approach. This design choice allows for a more equitable comparison, as models requiring only textual prompts or those designed for unrelated tasks (e.g., vector sketch generation or appearance transfer methods such as \cite{cia}) are fundamentally different in their objectives and are excluded from the subsequent visualizations. 
 \begin{figure*}
    \centering
    \includegraphics[width=0.95\linewidth]{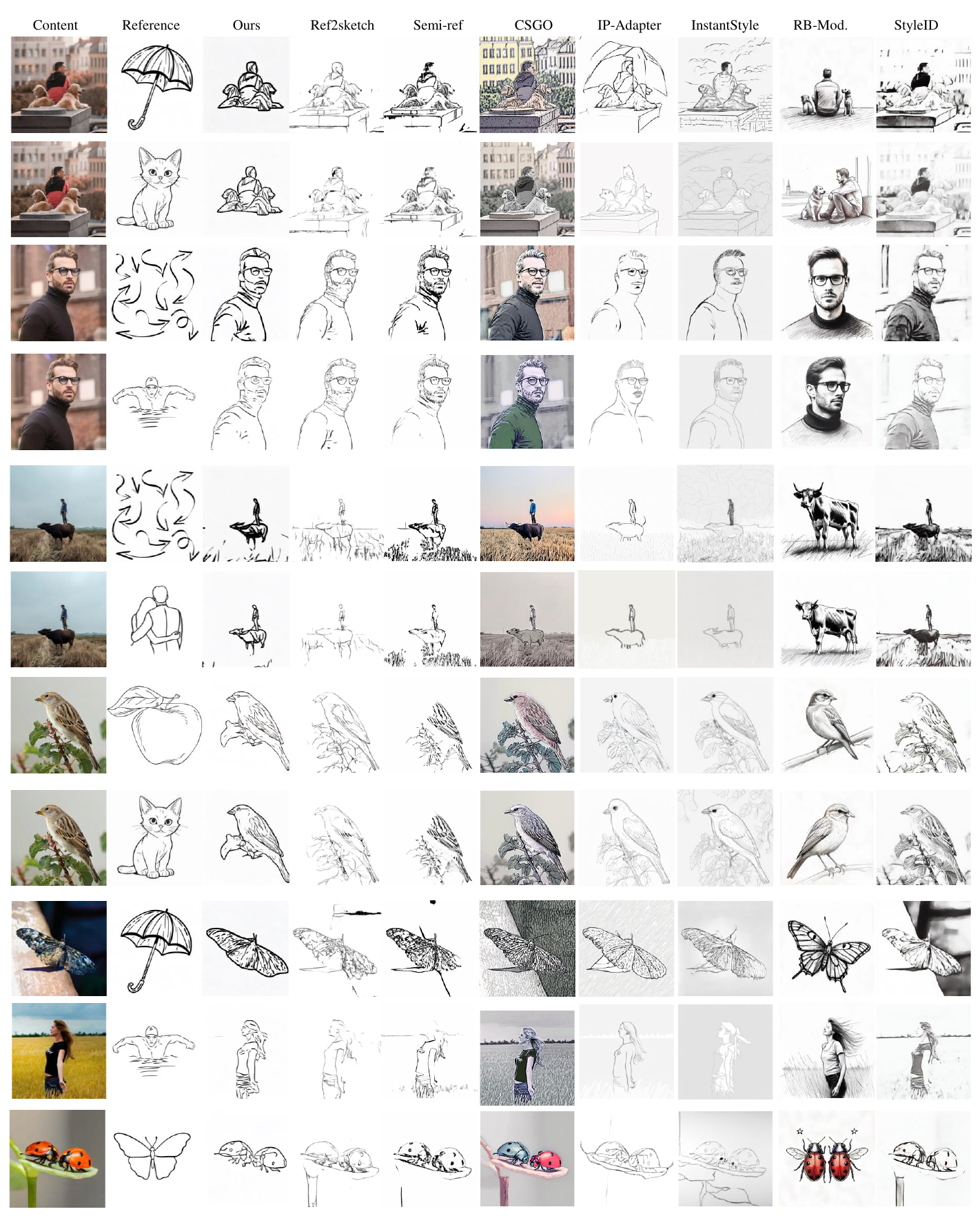}
 \caption{Comparison of sketches generated by Stroke2Sketch and baseline methods, including Ref2Sketch, Semi-ref2Sketch, CSGO, IP-Adapter, InstantStyle, RB-Modulation, and StyleID. Each row presents a content image, reference sketch, and results from different methods. Zoom in to view stroke details, highlighting the accurate alignment of stroke attributes and content semantics achieved by our approach.}
  \label{fig17}
    %\vspace{-4mm}
\end{figure*}
 \begin{figure*}
    \centering
    \includegraphics[width=0.95\linewidth]{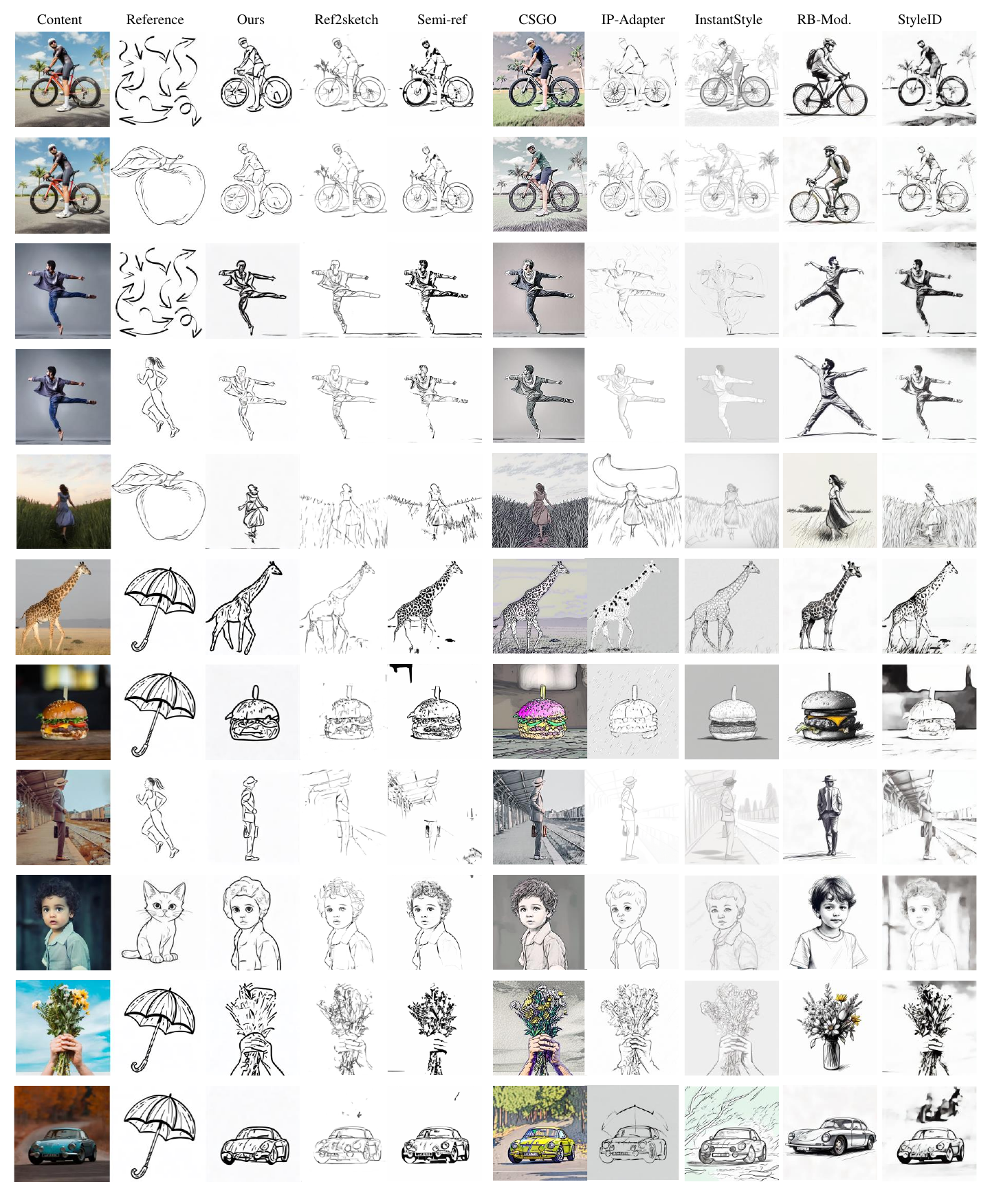}
  \caption{Additional qualitative comparison of Stroke2Sketch against baseline methods. The rows showcase content images, reference sketches, and outputs from various methods. Note the stroke details and style consistency in the results generated by our method. Zoom in to view stroke details for a clearer examination of stylistic fidelity and semantic alignment.}
  \label{fig18}
    %\vspace{-4mm}
\end{figure*}
 \begin{figure*}
    \centering
    \includegraphics[width=1\linewidth]{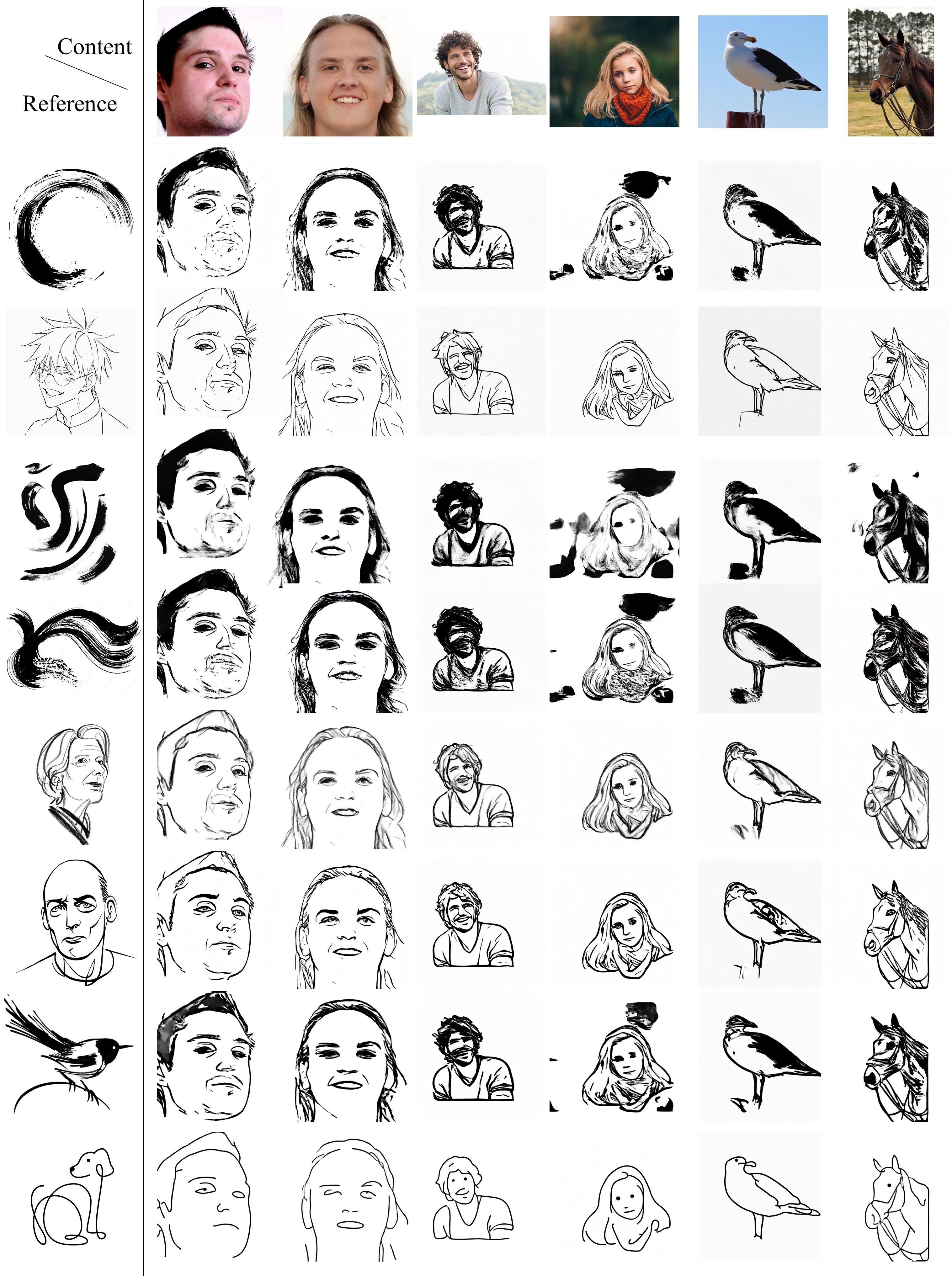}
\caption{Sketch generation results using Stroke2Sketch across diverse content and reference styles.}
  \label{fig19}
    %\vspace{-4mm}
\end{figure*}

\cref{fig17} and~\cref{fig18} present additional comparison results across diverse styles and content images, demonstrating the robustness of our method. Meanwhile, ~\cref{fig19} showcase sketches generated by Stroke2Sketch across different datasets, further validating its adaptability to varied styles and semantic requirements.

This focused evaluation highlights the advantages of our approach in achieving consistent stroke fidelity and semantic alignment while excluding comparisons with methods that do not align with the reference-based sketch extraction task.

\section{Failure Cases}\label{sec:d}
While our method demonstrates strong performance across a variety of reference styles, certain limitations remain when handling reference sketches with extreme characteristics. Specifically, sketches with overly simplistic or highly complex strokes pose challenges. As illustrated in ~\cref{fig19}, cases involving highly abstract continuous single-line references or densely detailed brushstroke references often result in suboptimal outcomes. 

For instance, overly thick or abstract strokes can lead to detail loss or distortions in features like facial expressions, particularly in areas such as the eyes or intricate textures. Similarly, when the reference sketch exhibits densely packed details, the model may struggle to balance semantic consistency and stroke fidelity, resulting in either excessive abstraction or loss of critical content elements. 

This behavior mimics how human artists adapt their interpretations based on the nature of the reference strokes. However, the challenge of fully decoupling semantic information from stroke attributes while maintaining both fidelity and style remains an open problem. Future work could explore advanced segmentation or attention mechanisms to address these limitations and enhance robustness in extreme cases.